\documentclass[10pt,twocolumn,letterpaper]{article}
\newcommand{\paperTitle}{Learning Dual-Continuous-Time Schedules for Diffusion Caching}
\newcommand{\paperAuthorBlock}{%
  Ziyu Li$^{1,}$\thanks{Equal contribution.}\quad
  Zhanyang Jin$^{2,}$\footnotemark[1]\quad
  Yingxin Xie$^{1}$\quad
  Shengchao Liu$^{1,}$\thanks{Corresponding author.}\\
  $^{1}$The Chinese University of Hong Kong\\
  $^{2}$University of Illinois Urbana-Champaign%
}
\newcommand{\paperAuthorMetadata}{Ziyu Li, Zhanyang Jin, Yingxin Xie, Shengchao Liu}
\date{}

\usepackage{times}
\usepackage[dvipsnames]{xcolor}
\usepackage{graphicx}
\usepackage{amsmath,amssymb}
\usepackage{booktabs}
\usepackage[numbers,sort&compress]{natbib}
\usepackage[format=plain,labelformat=simple,labelsep=period,font=small,compatibility=false]{caption}
\usepackage[hyphens]{url}
\usepackage{enumitem}
\setlist[itemize]{noitemsep,leftmargin=*,topsep=0pt}
\makeatletter
\renewcommand{\@maketitle}{%
  \newpage\null\vskip 0.375in
  \begin{center}
    {\Large\bfseries \@title\par}
    \vspace*{24pt}%
    {\large\setlength{\lineskip}{0.5em}%
      \begin{tabular}[t]{c}\@author\end{tabular}\par}
    \vskip 0.5em
    \vspace*{12pt}
  \end{center}}

\renewenvironment{abstract}
  {\centerline{\large\bfseries \abstractname}%
   \vspace*{12pt}\noindent\itshape\ignorespaces}
  {\vspace*{12pt}}

\NewDocumentCommand{\preprint@heading}{m s o m}{%
  \IfBooleanTF{#2}{#1*{#4}}{%
    \IfNoValueTF{#3}
      {#1{\texorpdfstring{\hskip-1em.~}{}#4}}
      {#1[#3]{\texorpdfstring{\hskip-1em.~}{}#4}}}}
\font\preprint@subheadingfont=ptmb scaled 1100
\font\preprint@minorheadingfont=ptmb scaled 1000
\renewcommand{\section}{\preprint@heading{%
  \@startsection{section}{1}{0pt}{-10pt plus -2pt minus -2pt}{7pt}{\large\bfseries}}}
\renewcommand{\subsection}{\preprint@heading{%
  \@startsection{subsection}{2}{0pt}{-8pt plus -2pt minus -2pt}{5pt}{\preprint@subheadingfont}}}
\renewcommand{\subsubsection}{\preprint@heading{%
  \@startsection{subsubsection}{3}{0pt}{-6pt plus -2pt minus -2pt}{3pt}{\preprint@minorheadingfont}}}
\makeatother

\usepackage{colortbl}
\usepackage{multirow}
\usepackage{algorithm}
\usepackage{algpseudocode}
\usepackage{placeins}
\usepackage{needspace}

\definecolor{linkblue}{rgb}{0.21,0.49,0.74}
\usepackage[pagebackref,breaklinks,colorlinks,allcolors=linkblue]{hyperref}
\hypersetup{pdfauthor={\paperAuthorMetadata}}
\usepackage[capitalize]{cleveref}
\crefname{section}{Sec.}{Secs.}
\Crefname{section}{Section}{Sections}
\crefname{table}{Tab.}{Tabs.}
\Crefname{table}{Table}{Tables}

\title{\paperTitle}
\author{\paperAuthorBlock}
\hypersetup{pdftitle={\paperTitle}}

\begin{document}
\maketitle
\begin{abstract}
Diffusion models generate high-quality data through an iterative denoising process, but the sequential computations incur substantial inference latency. Feature caching addresses this problem by reusing earlier features to approximate later ones, thus significantly accelerating generation. However, existing feature caching methods typically either make binary decisions at each timestep about whether to reuse cached features or adjust computation times while keeping the operation order fixed. Their limitations include insensitivity to trajectory geometry and the propagation of local approximation errors during feature reuse. To address these limitations, we propose to reformulate cache scheduling as a dual-continuous-time optimization problem, with two separate timelines. One timeline determines when to compute features, while the other determines when to reuse previously computed features, inspired by fast- and slow-varying denoising dynamics, respectively. We then merge these two timelines into a unified sampling schedule. More concretely, we train a stochastic policy via reinforcement learning to learn the positions on both timelines jointly, guided by global generation feedback. On SD3 Medium and SD3.5 Large with zero- and first-order caching at two computation budgets, our method outperforms four strong baselines in reference-based LPIPS and SSIM across all eight settings, while achieving $1.33$--$2.62\times$ speedups over full computation. Further trajectory analysis shows that the learned schedules reduce mean cumulative step-weighted local error by $14.7\%$--$79.4\%$.
\end{abstract}

\begin{figure*}[t]
  \centering
  \includegraphics[width=\textwidth]{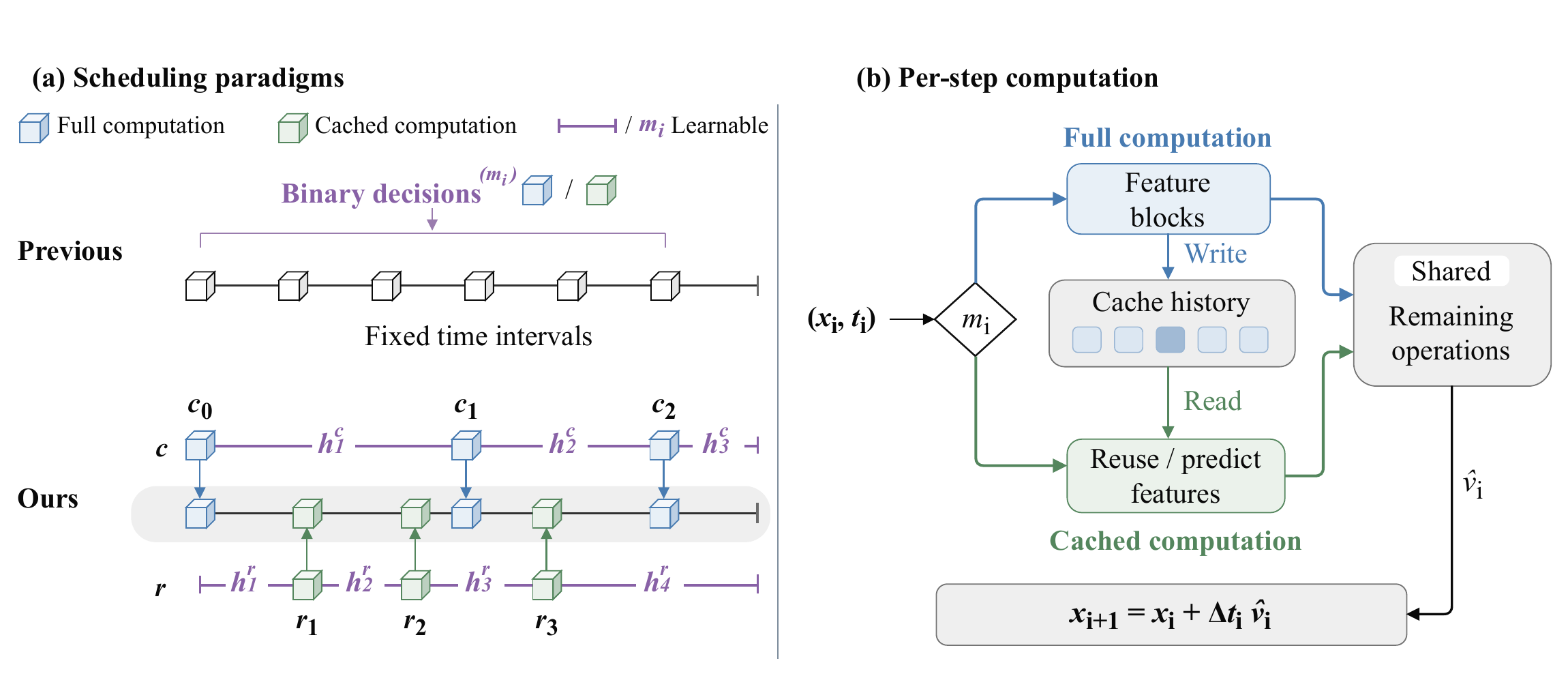}
  \input{figures/main-figure-caption}
\end{figure*}

\section{Introduction}
\label{sec:intro}

Diffusion~\citep{NEURIPS2020_4c5bcfec,song2021denoising,pmlr-v139-nichol21a,NEURIPS2021_49ad23d1}
and flow matching models~\citep{lipman2023flow,liu2023flow} enable high-quality
synthesis across
images~\citep{9878449,NEURIPS2022_ec795aea,10377858,ma2024sit},
videos~\citep{NEURIPS2022_39235c56,10203078}, and 3D
shapes~\citep{9578791,poole2023dreamfusion}, but sampling requires iterative
numerical integration~\citep{song2021score} with sequential network evaluations,
incurring substantial latency. Alongside fast solvers~\citep{NEURIPS2022_260a14ac,NEURIPS2023_9c2aa1e4},
distillation~\citep{salimans2022progressive,10655942}, consistency
training~\citep{pmlr-v202-song23a}, and parallel sampling~\citep{11515172},
feature caching exploits the slow variation of intermediate features across adjacent
steps. By storing features computed at full-computation steps for reuse or prediction at
later steps, it reduces inference cost while keeping the pretrained network’s weights
fixed~\citep{10655100,selvaraju2024forafastforwardcachingdiffusion,11444758}.

Given a feature reuse or prediction rule, the sampler needs to decide when to compute or
approximate features, a task known as cache scheduling. These decisions are tightly
coupled with numerical integration. Specifically, for a given cached prediction error, a
larger integration step causes a greater state deviation, affecting subsequent
predictions. Meanwhile, this error also depends on the available feature history.
Full-computation times determine which features are stored, while moving a cached step
across a full step changes the history used for feature prediction. Despite this
coupling, existing schedulers fix either the time
grid~\citep{aliev2026recache,Chung_2026_CVPR,bu2026dicache} or the operation
order~\citep{lei2026budgetconstrained}. In effect, a fixed grid restricts how integration intervals
are allocated to cached predictions with different approximation errors, whereas a fixed
order keeps the full computations preceding each cached step unchanged, even when their
times are adjusted. Together, these constraints limit the joint adaptation of local
prediction accuracy and its cumulative influence on the trajectory. We therefore learn
full- and cached-computation times under an exact budget, allowing step sizes and
operation order to adapt together.

To learn these times, we introduce a dual-continuous-time parameterization for cache
scheduling (Fig.~\ref{fig:joint-continuous-overview}). Specifically, we represent full- and
cached-computation times on two separate timelines and merge them into a unified
schedule. Fixing point counts on each timeline preserves an exact full-computation
budget, while shifting points adjusts integration step sizes and operation interleaving.
However, when points from the two timelines cross, the operation order switches
discretely, altering which features are available for reuse or prediction. Such discrete
switches make gradient-based optimization difficult, and backpropagating through
multi-step sampling is expensive. We therefore train a stochastic policy via
reinforcement learning, guided by final-image fidelity and preference feedback
(Fig.~\ref{fig:policy-training-overview}). The pretrained diffusion model remains frozen
during training, while inference applies a single deterministic schedule per setting
without runtime policy evaluation.

We evaluate our method on SD3 Medium~\citep{esser2024scaling} and SD3.5
Large~\citep{stabilityai2024sd35} with zero- and first-order caching under two
computation budgets. Across all eight settings, our method achieves lower LPIPS and
higher SSIM than four strong caching baselines at comparable inference latency,
delivering $1.33$--$2.62\times$ speedups over full 28-step sampling. Under the tighter
budget of 7 full-computation steps, it outperforms all baselines across all six fidelity
and preference metrics, reducing LPIPS by up to $38.7\%$ over the strongest
baseline~\citep{aliev2026recache}. Ablations confirm the benefit of joint time-and-order
optimization, while trajectory analysis shows that the learned schedules adapt
integration intervals to local approximation errors and achieve lower cumulative
step-weighted local error than uniform scheduling.

\begingroup
\setlength{\parskip}{0pt}
Our main contributions are summarized as follows:
\begin{itemize}
  \item We introduce a dual-continuous-time parameterization of full- and cached-computation
    times that allows integration step sizes and operation order to adapt together under an
    exact budget.
  \item We optimize this parameterization via reinforcement learning with final-image fidelity
    and human preference feedback, while keeping the pretrained generator frozen.
  \item Our method achieves the lowest LPIPS and highest SSIM across all eight evaluated
    settings at comparable latency. Under tight budgets, trajectory analyses show lower
    cumulative step-weighted local error than uniform scheduling despite higher mean local
    error.
\end{itemize}
\endgroup

\pagebreak[0]
\begin{samepage}
\section{Related work}
\label{sec:related}

\paragraph{Sampling schedules and solvers.}
Uncached samplers optimize time grids via solver-specific design~\citep{sabour2024align,10657720}, trajectory selection~\citep{chen2024on}, or differentiation~\citep{ICLR2025_75004615}. Relatedly, EPD-Solver learns continuous solver parameters with reinforcement learning~\citep{11515172}, while Bespoke solvers train model-specific integrators~\citep{shaul2024bespoke}. Whereas these methods improve numerical integration without feature reuse, reward-driven approaches such as DDPO~\citep{black2024training} and DRaFT~\citep{clark2024directly} fine-tune generator weights from reward feedback; in contrast, we keep the generator frozen.
\par
\end{samepage}

\paragraph{Feature reuse and prediction.}
DeepCache and Faster Diffusion reuse U-Net features~\citep{10655100,NEURIPS2024_9ad996b5}, whereas FORA caches attention and MLP outputs~\citep{selvaraju2024forafastforwardcachingdiffusion}. Feature reuse can also be controlled per layer~\citep{ma2024learningtocache} or per token~\citep{zou2025accelerating}, while TaylorSeer and HiCache forecast features from multi-step history~\citep{11444758,ICLR2026_b67c77f8}.

\paragraph{Cache scheduling on fixed time grids.}
Fixed-grid schedulers determine refreshes using timestep embeddings, spectral evolution, sensitivity estimates, or online probes~\citep{11093063,Chung_2026_CVPR,Haghighi_2026_CVPR,bu2026dicache}; along similar lines, AdaCache adapts refreshes to per-video residual changes~\citep{11445977}, and BAG conditions gating on a target budget~\citep{zhao2026bagbudgetawaregatingdiffusion}. Alternatively, search-based methods optimize discrete schedules~\citep{Cui_2026_CVPR,NEURIPS2025_bfa0d4d1,aggarwal2026evolutionary}, and ReCache learns binary masks from final-image feedback~\citep{aliev2026recache}. OTCache interpolates continuous warps across budgets before rounding and projecting them onto discrete step indices~\citep{gao2026otcache}.

\paragraph{Time adaptation for cached sampling.}
Beyond fixed grids, ERTACache analyzes feature-shift and step-amplification errors, combining timestep adjustment with explicit error rectification~\citep{peng2026ertacache}. BudCache first searches for a budget-constrained cache mask and then optimizes sampling times with that mask fixed~\citep{lei2026budgetconstrained}, while RAPID$^3$ learns state-dependent policies for timestep jumps, cache reuse, and sparse attention with a frozen generator~\citep{zhao2026rapid}. In contrast, we parameterize full and cached computations with separate interval families to learn their times jointly under an exact full-computation budget, allowing their relative order to adapt while keeping the caching rules fixed. We train this interval policy from final-image feedback and decode a shared deterministic schedule per setting offline, without online policy evaluation.

\section{Preliminaries}
\label{sec:preliminary}

\paragraph{Deterministic generative sampling.}
Diffusion~\citep{song2021score} and flow matching
models~\citep{lipman2023flow,liu2023flow} admit deterministic ODE sampling.
For a pretrained rectified-flow transformer~\citep{esser2024scaling}, the latent
state evolves as
\begin{equation}
 \frac{\mathrm d x(t)}{\mathrm d t}
 =v_\theta(x(t),t,c),\qquad x(t_{\mathrm{start}})=z,
\label{eq:generative_ode}
\end{equation}
where $v_\theta$ denotes the frozen velocity field, $z$ the initial noise, and $c$
the condition. As the noise level $t$ decreases toward the data endpoint along a
grid $t_0=t_{\mathrm{start}}>t_1>\cdots>t_N$, full-computation Euler sampling
starts from $x_0=z$ and iterates
\begin{equation}
 x_{i+1}=x_i+(t_{i+1}-t_i)v_\theta(x_i,t_i,c),
\label{eq:full_euler}
\end{equation}
after which a fixed decoder maps $x_N$ to the image $y=\operatorname{Dec}(x_N)$.

\paragraph{Cached sampling.}
To reduce inference cost, full computations evaluate the feature blocks and refresh
the cache history $\mathcal H_i$, whereas cached computations approximate selected
block outputs from $\mathcal H_i$. The shared remaining layers combine these
features with the current state and time to produce
$\widetilde v_\theta(x_i,t_i,c;\mathcal H_i)$, so even constant cached features can
yield distinct velocity estimates across steps.

Zero-order reuse copies the latest features. Given $F_a,F_b$ at the two latest
full-computation noise levels $t_a>t_b$, first-order linear extrapolation to
$t_i<t_b$ gives
\begin{equation}
 \widehat F_i=F_b+\frac{t_b-t_i}{t_a-t_b}(F_b-F_a),
\label{eq:sigma_feature_forecast}
\end{equation}
reverting to zero-order reuse when only one prior full computation is available.

\section{Method}
\label{sec:method}

We learn full- and cached-computation times from final-image feedback using the two-family construction in \cref{fig:joint-continuous-overview} and the training procedure in \cref{fig:policy-training-overview}.

\begin{figure*}[t]
  \centering
  \includegraphics[width=\textwidth]{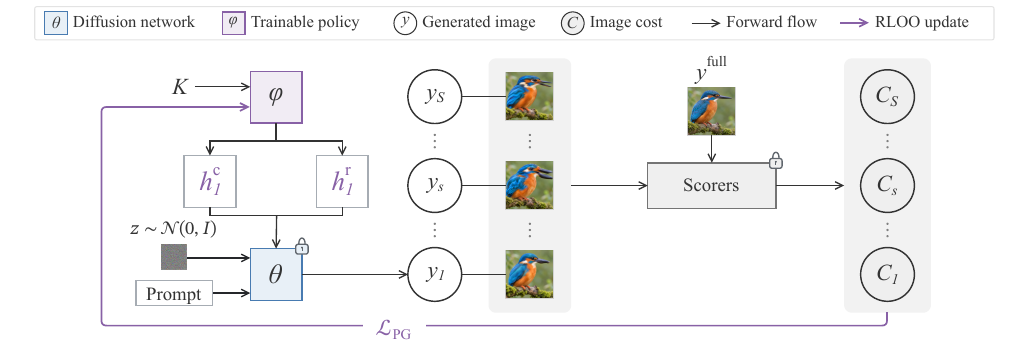}
  \input{figures/policy-training-caption}
\end{figure*}

\subsection{Budgeted cached integration}
\label{sec:method_setup}

Consider $N$ solver updates, of which $K\in\{1,\ldots,N\}$ perform full computations. Normalized progress $u\in[0,1]$ increases from initial noise ($u=0$) to the final sample ($u=1$), and a fixed monotone map $t=\psi(u)$ converts progress to solver time, with $u_i=i/N$ reproducing the native grid.

A schedule comprises boundaries $0=u_0<\cdots<u_N=1$ and operation labels $m_i\in\{0,1\}$ for $i=0,\ldots,N-1$. At $t_i=\psi(u_i)$, the cached counterpart of \cref{eq:full_euler} iterates
\begin{equation}
\begin{split}
 \widehat v_i &=
 \begin{cases}
 v_\theta(x_i,t_i,c), & m_i=1,\\
 \widetilde v_\theta(x_i,t_i,c;\mathcal H_i), & m_i=0,
 \end{cases}\\
 x_{i+1} &= x_i+(t_{i+1}-t_i)\widehat v_i ,
\end{split}
\label{eq:cached_euler}
\end{equation}
where full computations refresh $\mathcal H_i$ and $\widetilde v_\theta$ applies
the selected cache backend, with $m_0=1$ and $\sum_{i=0}^{N-1}m_i=K$.

In \cref{eq:cached_euler}, shifting computation times alters both velocity
estimates and integration weights, whereas the operation order governs the
available cache history. We therefore jointly learn full- and
cached-computation times, which together determine the integration grid and
operation order.

\subsection{Dual-continuous-time parameterization}
\label{sec:method_points}

Anchoring the initial full computation at $c_0=0$ partitions $[0,1]$ into $K$
intervals: $K-1$ inter-refresh gaps and the final gap to $1$. Similarly,
placing $N-K$ cached-computation points defines $N-K+1$ intervals, including
both endpoint gaps. To move these points while preserving the exact budget and
within-family order, we represent the two timelines via positive, normalized
interval vectors $h^{\mathrm c}\in\Delta^{K-1}_{+}$ and
$h^{\mathrm r}\in\Delta^{N-K}_{+}$ on the open simplex
$\Delta^{d-1}_{+}=\{h\in\mathbb R^d:h_j>0,\ \sum_j h_j=1\}$.
Cumulative sums recover the point locations:
\begin{subequations}
\label{eq:interval_simplices}
\begin{align}
 c_0=0,\quad c_j&=\sum_{\ell=1}^{j}h^{\mathrm c}_\ell
 \quad (j=1,\ldots,K-1),
\label{eq:compute_points}
 \\
 r_j&=\sum_{\ell=1}^{j}h^{\mathrm r}_\ell
 \quad (j=1,\ldots,N-K).
\label{eq:reuse_points}
\end{align}
\end{subequations}
Sorting labeled points by progress yields the schedule:
\begin{equation}
\begin{split}
 &\bigl((u_i,m_i)\bigr)_{i=0}^{N-1}\\
 &\quad=\operatorname{sort}_u\!\left(
 \{(c_j,1)\}_{j=0}^{K-1}\cup\{(r_j,0)\}_{j=1}^{N-K}\right),
\end{split}
\label{eq:labeled_schedule}
\end{equation}
with the terminal boundary $u_N=1$ adding no computation.

This parameterization provides $(K-1)+(N-K)=N-1$ continuous degrees of freedom
under an exact budget of $K$ full computations. While positive intervals
preserve within-family order, cross-family crossings reorder operations and
update the available cache history. Mapping the merged points through $\psi$
yields solver times whose adjacent differences specify the Euler step sizes.
In first-order forecasting (\cref{eq:sigma_feature_forecast}), the history
spacing $t_a-t_b$ and extrapolation distance $t_b-t_i$ jointly set the
forecasting ratio, while the resulting velocity estimate is scaled by the
integration interval $t_{i+1}-t_i$ in \cref{eq:cached_euler}.

\subsection{Interval policy}
\label{sec:policy}

Because time-point crossings switch operation order and cache dependencies
discretely, we optimize the intervals from final-image feedback via a
stochastic policy $p_\phi(h\mid K)$ over $h=(h^{\mathrm c},h^{\mathrm r})$,
avoiding differentiation through sorting or multi-step generation. Specifically,
we train a dedicated policy for each model, caching backend, and budget $K$.
Applying a fixed invertible log-ratio transform
$T$~\citep{19b3d08d-3e3e-3ca5-a174-44ff933afefb} yields unconstrained coordinates
$q=T(h^{\mathrm c},h^{\mathrm r})\in\mathbb{R}^{N-1}$, allowing Gaussian
samples to map back through $T^{-1}$ to positive, normalized intervals
(density and coordinate details in Supplementary Secs.~A and B).

Let $\bar h$ denote the two interval vectors extracted from a uniform
reference schedule. We parameterize the policy center as
$\mu_\phi(K)=T(\bar h)+f_\phi(K)$, learning a coordinate-space residual around
the uniform reference. With positive exploration scales $s_\ell$ at update
$\ell$, we sample
\begin{equation}
  q = \mu_\phi(K) + s_\ell \odot \epsilon,
  \quad
  \epsilon \sim \mathcal{N}(0, I),
  \quad
  h = T^{-1}(q),
  \label{eq:gaussian_sampling}
\end{equation}
and decode each sample $h$ into a labeled grid via
\cref{eq:interval_simplices,eq:labeled_schedule}. Although coordinate
perturbations are drawn independently for the two families, both are coupled
through the shared final-image cost.

\subsection{Learning from completed generations}
\label{sec:method_training}

For a condition--noise pair $(c,z)$, let $y(h;c,z)$ denote the image generated
under $h$ and $y^{\mathrm{full}}(c,z)$ its reference on the native $N$-step
grid. Following ReCache~\citep{aliev2026recache}, we minimize
\begin{equation}
\begin{split}
 C(h;c,z)
 &=D_{\mathrm{patch}}\!\left(y(h;c,z),y^{\mathrm{full}}(c,z)\right)\\
 &\quad-\alpha R_{\mathrm{train}}\!\left(y(h;c,z),c\right),\\
 J(\phi;K)&=\mathbb E_{(c,z)\sim\mathcal D}
             \mathbb E_{h\sim p_\phi(\cdot\mid K)}[C(h;c,z)] ,
\end{split}
\label{eq:schedule_objective}
\end{equation}
where $D_{\mathrm{patch}}$ is patchwise LPIPS~\citep{8578166}
and $R_{\mathrm{train}}$ is HPSv2~\citep{wu2023human} weighted by $\alpha$.

We treat schedule decoding ($\operatorname{sort}_u$), ODE integration, and
image scoring as a black-box environment. For each pair $(c,z)$, we draw
$S\geq2$ candidate schedules via \cref{eq:gaussian_sampling} and generate each
image with its own cache history. Using the RLOO
baseline~\citep{williams1992reinforce,kool2019buy,ahmadian-etal-2024-back,aliev2026recache}
$b_s=(S-1)^{-1}\sum_{k\ne s}C_k$, we estimate the policy gradient with the
score-function estimator~\citep{NIPS2015_de03beff}:
\begin{equation}
 \widehat{\nabla_\phi J}
 =\frac{1}{S}\sum_{s=1}^{S}
 (C_s-b_s)\nabla_\phi\log p_\phi(h_s\mid K).
\label{eq:rloo_gradient}
\end{equation}
Closed-form Gaussian score gradients are evaluated at the fixed sampled
intervals while keeping the generator, caching rules, and scorers frozen.
Because the simplex parameterization enforces the budget exactly, no penalty
term is needed (derivations and algorithms in Supplementary Sec.~A).

\subsection{Deterministic inference}
\label{sec:method_inference}

At inference, we deterministically decode the policy center
$h^\star=T^{-1}(\mu_\phi(K))$ via
\cref{eq:interval_simplices,eq:labeled_schedule}. This schedule is precomputed
offline and shared across prompts and latents within each setting, executing
$K$ full and $N-K$ cached computations with zero scheduling overhead.

\ifdefined\table
  \let\endMainResultsPreview\relax
\else
  \documentclass[10pt,letterpaper]{article}
  \usepackage[textwidth=6.875in,vmargin=0.75in]{geometry}
  \usepackage{booktabs,times,xcolor,colortbl,multirow}
  \pagestyle{empty}
  \def\endMainResultsPreview{\end{document}}
  \begin{document}
\fi
\definecolor{mainResultCyan}{rgb}{0.925,1,1}
\begin{table*}[t]
\centering
\caption{\textbf{Scheduling comparisons on SD3 Medium.}
We evaluate zero- and first-order feature caching over $N=28$ steps at target
budgets of $K=7/14$ full computations across 1,000 DrawBench images (PSNR in dB).
Bold and underline mark the best and second-best scores per backend and budget.}
\label{tab:sd3_experiments}
\begingroup
\fontsize{9}{10.8}\selectfont
\setlength{\tabcolsep}{3.2pt}
\renewcommand{\arraystretch}{1.12}
\textbf{(a) Zero-order}\par
\vspace{3pt}
\begin{tabular}{c*{12}{c}}
\toprule
\multirow{2}{*}{\textbf{Scheduling policy}}
& \multicolumn{2}{c}{\textbf{LPIPS\boldmath$\downarrow$}}
& \multicolumn{2}{c}{\textbf{SSIM\boldmath$\uparrow$}}
& \multicolumn{2}{c}{\textbf{PSNR\boldmath$\uparrow$}}
& \multicolumn{2}{c}{\textbf{ImageReward\boldmath$\uparrow$}}
& \multicolumn{2}{c}{\textbf{PickScore\boldmath$\uparrow$}}
& \multicolumn{2}{c}{\textbf{HPSv2\boldmath$\uparrow$}}
\\
\cmidrule(lr){2-3}\cmidrule(lr){4-5}\cmidrule(lr){6-7}\cmidrule(lr){8-9}\cmidrule(lr){10-11}\cmidrule(lr){12-13}
& \makebox[28pt]{\textbf{\boldmath$K{=}7$}}
& \makebox[28pt]{\textbf{\boldmath$K{=}14$}}
& \makebox[28pt]{\textbf{\boldmath$K{=}7$}}
& \makebox[28pt]{\textbf{\boldmath$K{=}14$}}
& \makebox[28pt]{\textbf{\boldmath$K{=}7$}}
& \makebox[28pt]{\textbf{\boldmath$K{=}14$}}
& \makebox[28pt]{\textbf{\boldmath$K{=}7$}}
& \makebox[28pt]{\textbf{\boldmath$K{=}14$}}
& \makebox[28pt]{\textbf{\boldmath$K{=}7$}}
& \makebox[28pt]{\textbf{\boldmath$K{=}14$}}
& \makebox[28pt]{\textbf{\boldmath$K{=}7$}}
& \makebox[28pt]{\textbf{\boldmath$K{=}14$}}
\\
\midrule
Uniform & 0.6804 & 0.5430 & 0.3994 & 0.4739 & 11.67 & 12.08 & -1.307 & 0.033 & 19.65 & 21.00 & 0.2428 & 0.2690 \\
DiCache & 0.4073 & 0.1593 & 0.4392 & 0.8059 & 14.78 & 20.39 & -0.122 & 0.517 & 20.36 & 21.52 & 0.2642 & 0.2804 \\
SeaCache & 0.4520 & 0.1516 & 0.4828 & 0.8125 & 13.24 & 21.00 & -0.119 & 0.530 & 20.52 & 21.53 & 0.2662 & 0.2807 \\
ReCache & \underline{0.2815} & \underline{0.1433} & \underline{0.5878} & \underline{0.8243} & \underline{16.92} & \underline{21.13} & \underline{0.272} & \underline{0.531} & \underline{21.00} & \textbf{21.58} & \underline{0.2734} & \textbf{0.2810} \\
\rowcolor{mainResultCyan}
\textbf{Ours} & \textbf{0.2678} & \textbf{0.1326} & \textbf{0.6746} & \textbf{0.8307} & \textbf{17.38} & \textbf{21.29} & \textbf{0.298} & \textbf{0.533} & \textbf{21.17} & \underline{21.54} & \textbf{0.2759} & \underline{0.2808} \\
\bottomrule
\end{tabular}\par
\vspace{10pt}
\textbf{(b) First-order}\par
\vspace{3pt}
\begin{tabular}{c*{12}{c}}
\toprule
\multirow{2}{*}{\textbf{Scheduling policy}}
& \multicolumn{2}{c}{\textbf{LPIPS\boldmath$\downarrow$}}
& \multicolumn{2}{c}{\textbf{SSIM\boldmath$\uparrow$}}
& \multicolumn{2}{c}{\textbf{PSNR\boldmath$\uparrow$}}
& \multicolumn{2}{c}{\textbf{ImageReward\boldmath$\uparrow$}}
& \multicolumn{2}{c}{\textbf{PickScore\boldmath$\uparrow$}}
& \multicolumn{2}{c}{\textbf{HPSv2\boldmath$\uparrow$}}
\\
\cmidrule(lr){2-3}\cmidrule(lr){4-5}\cmidrule(lr){6-7}\cmidrule(lr){8-9}\cmidrule(lr){10-11}\cmidrule(lr){12-13}
& \makebox[28pt]{\textbf{\boldmath$K{=}7$}}
& \makebox[28pt]{\textbf{\boldmath$K{=}14$}}
& \makebox[28pt]{\textbf{\boldmath$K{=}7$}}
& \makebox[28pt]{\textbf{\boldmath$K{=}14$}}
& \makebox[28pt]{\textbf{\boldmath$K{=}7$}}
& \makebox[28pt]{\textbf{\boldmath$K{=}14$}}
& \makebox[28pt]{\textbf{\boldmath$K{=}7$}}
& \makebox[28pt]{\textbf{\boldmath$K{=}14$}}
& \makebox[28pt]{\textbf{\boldmath$K{=}7$}}
& \makebox[28pt]{\textbf{\boldmath$K{=}14$}}
& \makebox[28pt]{\textbf{\boldmath$K{=}7$}}
& \makebox[28pt]{\textbf{\boldmath$K{=}14$}}
\\
\midrule
Uniform & 0.6526 & 0.5421 & 0.4281 & 0.4805 & 11.71 & 11.95 & -1.118 & -0.006 & 19.81 & 20.92 & 0.2471 & 0.2678 \\
DiCache & 0.6242 & 0.1735 & 0.2908 & 0.7903 & 9.92 & 19.24 & -0.884 & 0.484 & 19.54 & 21.45 & 0.2527 & 0.2797 \\
SeaCache & 0.5161 & \underline{0.1329} & 0.4454 & 0.8405 & 12.46 & 21.08 & -0.711 & \underline{0.568} & 19.80 & 21.58 & 0.2563 & \underline{0.2814} \\
ReCache & \underline{0.3905} & 0.1330 & \underline{0.5093} & \underline{0.8411} & \underline{14.26} & \underline{21.11} & \underline{-0.213} & \textbf{0.571} & \underline{20.19} & \underline{21.59} & \underline{0.2638} & \textbf{0.2815} \\
\rowcolor{mainResultCyan}
\textbf{Ours} & \textbf{0.2395} & \textbf{0.1328} & \textbf{0.7097} & \textbf{0.8412} & \textbf{17.62} & \textbf{21.15} & \textbf{0.350} & 0.548 & \textbf{21.18} & \textbf{21.60} & \textbf{0.2721} & 0.2810 \\
\bottomrule
\end{tabular}\par

\endgroup
\end{table*}
\begin{table*}[t]
\centering
\caption{\textbf{Scheduling comparisons on SD3.5 Large.}
We evaluate zero- and first-order feature caching over $N=28$ steps at target
budgets of $K=7/14$ full computations across 1,000 DrawBench images (PSNR in dB).
Bold and underline mark the best and second-best scores per backend and budget.}
\label{tab:sd35_experiments}
\begingroup
\fontsize{9}{10.8}\selectfont
\setlength{\tabcolsep}{3.2pt}
\renewcommand{\arraystretch}{1.12}
\textbf{(a) Zero-order}\par
\vspace{3pt}
\begin{tabular}{c*{12}{c}}
\toprule
\multirow{2}{*}{\textbf{Scheduling policy}}
& \multicolumn{2}{c}{\textbf{LPIPS\boldmath$\downarrow$}}
& \multicolumn{2}{c}{\textbf{SSIM\boldmath$\uparrow$}}
& \multicolumn{2}{c}{\textbf{PSNR\boldmath$\uparrow$}}
& \multicolumn{2}{c}{\textbf{ImageReward\boldmath$\uparrow$}}
& \multicolumn{2}{c}{\textbf{PickScore\boldmath$\uparrow$}}
& \multicolumn{2}{c}{\textbf{HPSv2\boldmath$\uparrow$}}
\\
\cmidrule(lr){2-3}\cmidrule(lr){4-5}\cmidrule(lr){6-7}\cmidrule(lr){8-9}\cmidrule(lr){10-11}\cmidrule(lr){12-13}
& \makebox[28pt]{\textbf{\boldmath$K{=}7$}}
& \makebox[28pt]{\textbf{\boldmath$K{=}14$}}
& \makebox[28pt]{\textbf{\boldmath$K{=}7$}}
& \makebox[28pt]{\textbf{\boldmath$K{=}14$}}
& \makebox[28pt]{\textbf{\boldmath$K{=}7$}}
& \makebox[28pt]{\textbf{\boldmath$K{=}14$}}
& \makebox[28pt]{\textbf{\boldmath$K{=}7$}}
& \makebox[28pt]{\textbf{\boldmath$K{=}14$}}
& \makebox[28pt]{\textbf{\boldmath$K{=}7$}}
& \makebox[28pt]{\textbf{\boldmath$K{=}14$}}
& \makebox[28pt]{\textbf{\boldmath$K{=}7$}}
& \makebox[28pt]{\textbf{\boldmath$K{=}14$}}
\\
\midrule
Uniform & 0.6213 & 0.4724 & 0.4212 & 0.5610 & 13.86 & 15.06 & -0.417 & 0.356 & 20.79 & 21.59 & 0.2606 & 0.2752 \\
DiCache & 0.4464 & 0.1475 & 0.3891 & \underline{0.8457} & 16.24 & 24.10 & -0.308 & 0.513 & 20.33 & 21.78 & 0.2590 & 0.2791 \\
SeaCache & 0.3922 & 0.1495 & 0.5618 & 0.8342 & 17.54 & 24.11 & 0.115 & 0.515 & 21.09 & 21.76 & 0.2692 & 0.2794 \\
ReCache & \underline{0.3175} & \underline{0.1349} & \underline{0.5680} & 0.8413 & \underline{19.18} & \textbf{24.26} & \underline{0.317} & \underline{0.540} & \underline{21.26} & \underline{21.79} & \underline{0.2716} & \textbf{0.2801} \\
\rowcolor{mainResultCyan}
\textbf{Ours} & \textbf{0.2916} & \textbf{0.1294} & \textbf{0.7024} & \textbf{0.8633} & \textbf{19.78} & \underline{24.12} & \textbf{0.323} & \textbf{0.543} & \textbf{21.47} & \textbf{21.82} & \textbf{0.2728} & \underline{0.2797} \\
\bottomrule
\end{tabular}\par
\vspace{10pt}
\textbf{(b) First-order}\par
\vspace{3pt}
\begin{tabular}{c*{12}{c}}
\toprule
\multirow{2}{*}{\textbf{Scheduling policy}}
& \multicolumn{2}{c}{\textbf{LPIPS\boldmath$\downarrow$}}
& \multicolumn{2}{c}{\textbf{SSIM\boldmath$\uparrow$}}
& \multicolumn{2}{c}{\textbf{PSNR\boldmath$\uparrow$}}
& \multicolumn{2}{c}{\textbf{ImageReward\boldmath$\uparrow$}}
& \multicolumn{2}{c}{\textbf{PickScore\boldmath$\uparrow$}}
& \multicolumn{2}{c}{\textbf{HPSv2\boldmath$\uparrow$}}
\\
\cmidrule(lr){2-3}\cmidrule(lr){4-5}\cmidrule(lr){6-7}\cmidrule(lr){8-9}\cmidrule(lr){10-11}\cmidrule(lr){12-13}
& \makebox[28pt]{\textbf{\boldmath$K{=}7$}}
& \makebox[28pt]{\textbf{\boldmath$K{=}14$}}
& \makebox[28pt]{\textbf{\boldmath$K{=}7$}}
& \makebox[28pt]{\textbf{\boldmath$K{=}14$}}
& \makebox[28pt]{\textbf{\boldmath$K{=}7$}}
& \makebox[28pt]{\textbf{\boldmath$K{=}14$}}
& \makebox[28pt]{\textbf{\boldmath$K{=}7$}}
& \makebox[28pt]{\textbf{\boldmath$K{=}14$}}
& \makebox[28pt]{\textbf{\boldmath$K{=}7$}}
& \makebox[28pt]{\textbf{\boldmath$K{=}14$}}
& \makebox[28pt]{\textbf{\boldmath$K{=}7$}}
& \makebox[28pt]{\textbf{\boldmath$K{=}14$}}
\\
\midrule
Uniform & 0.5825 & 0.4607 & 0.4680 & 0.5730 & 13.83 & 14.99 & -0.311 & 0.373 & 20.75 & 21.57 & 0.2629 & 0.2759 \\
DiCache & 0.5986 & 0.1411 & 0.3400 & 0.8522 & 11.40 & 23.29 & -0.807 & 0.542 & 19.61 & 21.77 & 0.2525 & 0.2799 \\
SeaCache & 0.4597 & 0.1057 & 0.5191 & \underline{0.8892} & 14.87 & 25.44 & -0.242 & 0.573 & 20.39 & 21.83 & 0.2630 & \underline{0.2805} \\
ReCache & \underline{0.3188} & \underline{0.1037} & \underline{0.6474} & 0.8890 & \underline{17.78} & \textbf{25.58} & \underline{0.276} & \textbf{0.582} & \underline{21.13} & \underline{21.84} & \underline{0.2727} & \textbf{0.2806} \\
\rowcolor{mainResultCyan}
\textbf{Ours} & \textbf{0.2779} & \textbf{0.1018} & \textbf{0.6845} & \textbf{0.8917} & \textbf{18.91} & \underline{25.46} & \textbf{0.300} & \underline{0.581} & \textbf{21.24} & \textbf{21.84} & \textbf{0.2734} & 0.2801 \\
\bottomrule
\end{tabular}\par

\endgroup
\end{table*}
\endMainResultsPreview
\begin{figure*}[t]
  \centering
  \includegraphics[width=\textwidth]{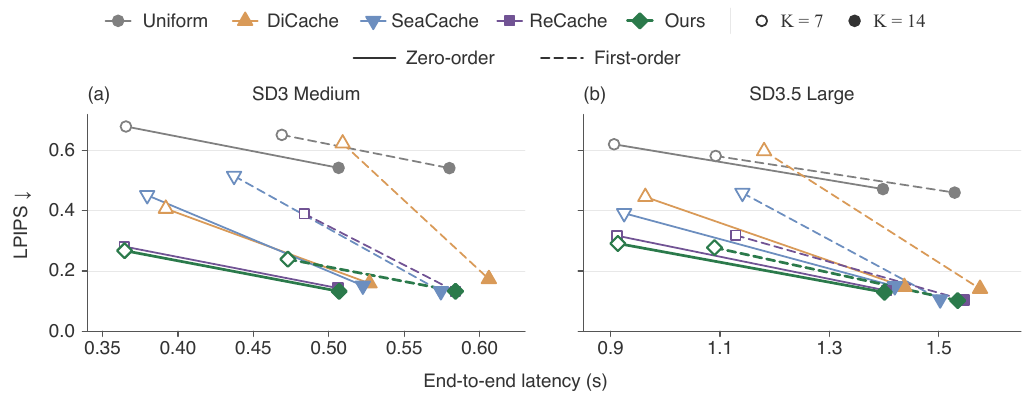}
  \caption{\textbf{Quality--latency trade-off.}
  Solid and dashed segments connect budgets $K=7$ (open markers) and $K=14$
  (filled markers) for zero- and first-order caching at $N=28$, with color and
  marker shape identifying each policy. Static policies apply a single
  deterministic schedule across runs, whereas SeaCache and DiCache make
  input-dependent refresh decisions. LPIPS is evaluated over the 1,000
  DrawBench images in \cref{tab:sd3_experiments,tab:sd35_experiments}, and mean
  end-to-end latency is averaged over 100 runs per setting (full timings and
  hardware details in Supplementary Sec.~C).}
  \label{fig:quality-latency}
\end{figure*}
\begin{figure*}[t]
  \centering
  \includegraphics[width=\textwidth]{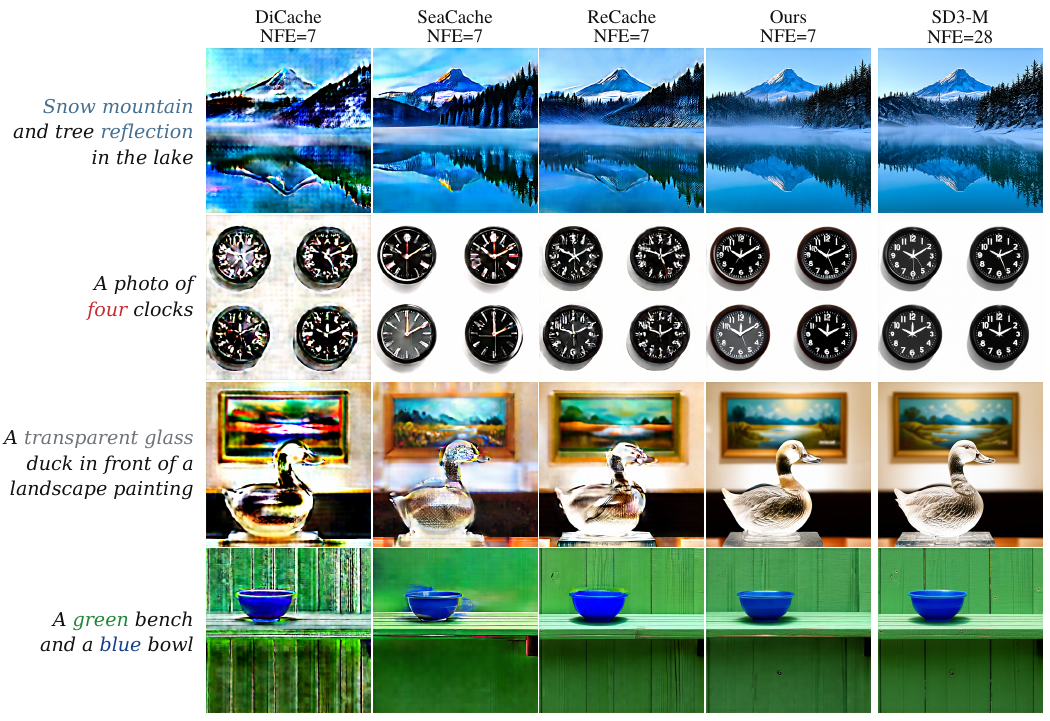}
  \caption{\textbf{Qualitative comparison with first-order feature forecasting.}
  Within each row, all methods share the same prompt and initial noise, with
  the rightmost column showing the uncached 28-step reference. NFE denotes the
  target budget of full-network computations (online refresh schedules for
  DiCache and SeaCache may vary across images).}
  \label{fig:qualitative-comparison}
\end{figure*}

\Needspace{5\baselineskip}
\section{Experiments}
\label{sec:experiments}

\subsection{Experimental setup}
\label{sec:experiments_setup}

\paragraph{Models and evaluation.}
We evaluate SD3 Medium~\citep{esser2024scaling} and SD3.5
Large~\citep{stabilityai2024sd35} at $512\times512$ resolution with $N=28$
Euler updates (shift $1$) on 200 DrawBench
prompts~\citep{NEURIPS2022_ec795aea} using five seeds each (1,000 images).
All methods share prompts, initial latents, and 28-step full-computation
references under budgets $K\in\{7,14\}$ (25\% and 50\% full computations). We
measure fidelity via whole-image LPIPS~\citep{8578166}
($\downarrow$), SSIM~\citep{1284395} ($\uparrow$), and PSNR ($\uparrow$),
and preference via ImageReward~\citep{NEURIPS2023_33646ef0},
PickScore~\citep{NEURIPS2023_73aacd8b}, and HPSv2~\citep{wu2023human} ($\uparrow$).

\paragraph{Backends and policies.}
We pair zero-order reuse (FORA-style)~\citep{selvaraju2024forafastforwardcachingdiffusion} and
first-order forecasting (TaylorSeer-style)~\citep{11444758}
(\cref{sec:preliminary,eq:sigma_feature_forecast}) with four fixed-grid
baselines: static Uniform (even spacing) and ReCache~\citep{aliev2026recache}
(learned positions), alongside adaptive SeaCache~\citep{Chung_2026_CVPR} and
DiCache~\citep{bu2026dicache} (using spectral-evolution and online-probe
criteria, with DiCache using the selected backend). Uniform, ReCache, and our
method enforce $K$ exactly, whereas SeaCache and DiCache target a mean of $K$.

\paragraph{Training and reporting.}
For each model, backend, and budget $K$, we train our policy on MS-COCO
captions~\citep{10.1007/978-3-319-10602-1_48} using the objective in
\cref{sec:method_training} ($S=6$, $\alpha=7$), and train a dedicated ReCache
policy for 1,000 updates under the same data and objective (for SD3 Medium at
$K=7$, both zero- and first-order policies are trained for 1,500 updates).
We evaluate each learned policy via its deterministic center schedule across
quality, latency, and trajectory analyses (experimental details and training
dynamics in Supplementary Secs.~\mbox{B--D}).

\subsection{Main results}
\label{sec:experiments_main}

\paragraph{Quantitative comparison.}
At comparable latency, our method achieves the lowest LPIPS and highest
SSIM across all eight model--backend--budget settings
(\cref{tab:sd3_experiments,tab:sd35_experiments}). Gains are largest at $K=7$
with first-order caching, reducing LPIPS over ReCache from $0.3905$ to
$0.2395$ (38.7\%) on SD3 Medium and from $0.3188$ to $0.2779$ (12.8\%) on SD3.5
Large. Across all four $K=7$ settings, our schedules outperform all baselines
on all six fidelity and preference metrics. At $K=14$, LPIPS reductions over
ReCache are more modest (0.2--7.5\%), while SSIM improves in every setting and
preference metrics show minor trade-offs across scorers.

\paragraph{Quality-latency trade-off.}
Requiring no online probing, our schedules achieve
$1.33$--$2.62\times$ speedups over full 28-step inference
(\cref{fig:quality-latency}), with mean latency differing from Uniform caching
by only $-0.25\%$ to $+0.86\%$.

\paragraph{Qualitative comparison.}
Under tight budgets (\cref{fig:qualitative-comparison}), fixed-grid baselines
produce irregular clock-face textures and distorted lake reflections, whereas
our first-order schedules reduce these artifacts and preserve the reference
geometry and fine contours more closely.

\subsection{Ablation and additional comparisons}
\label{sec:experiments_ablation}

\definecolor{mainResultCyan}{rgb}{0.925,1,1}
\begin{table}[!tbp]
\centering
\caption{\textbf{Continuous-time scheduling comparisons on SD3.5 Large ($N=28$).}
Full-computation timing fixes cached-computation times; the fixed-mask
controls fix the operation order. Joint learns both full- and cached-computation times, allowing their
interleaving to change. ReCache-mask uses 1,000 mask-learning and 500 time-learning
updates. Joint training ends at 1,500 updates for both zero-order budgets,
and at 912 and 1,000 updates for first-order $K=7$ and $K=14$, respectively.
Order 0/1 denotes zero-/first-order caching; bold and underline mark the
best and second-best scores per backend and budget.}
\label{tab:schedule_ablation}
\begingroup
\fontsize{9}{10.8}\selectfont
\setlength{\tabcolsep}{2pt}
\renewcommand{\arraystretch}{1.12}
\begin{tabular}{cc*{4}{c}}
\toprule
\multirow{2}{*}{\textbf{Order}}
&
\multirow{2}{*}{\textbf{Scheduling policy}}
& \multicolumn{2}{c}{\textbf{LPIPS\boldmath$\downarrow$}}
& \multicolumn{2}{c}{\textbf{ImageReward\boldmath$\uparrow$}} \\
\cmidrule(lr){3-4}\cmidrule(lr){5-6}
& & \makebox[28pt]{\textbf{\boldmath$K{=}7$}} & \makebox[28pt]{\textbf{\boldmath$K{=}14$}}
& \makebox[28pt]{\textbf{\boldmath$K{=}7$}} & \makebox[28pt]{\textbf{\boldmath$K{=}14$}} \\
\midrule
\multirow{4}{*}{0} & Full-comp. timing & \underline{0.3086} & \textbf{0.1172} & 0.313 & 0.529 \\
& Uniform-mask & 0.3214 & 0.1392 & 0.286 & \underline{0.538} \\
& ReCache-mask & 0.3087 & 0.1305 & \underline{0.320} & \textbf{0.543} \\
\rowcolor{mainResultCyan}
& \textbf{Joint} & \textbf{0.2916} & \underline{0.1294} & \textbf{0.323} & \textbf{0.543} \\
\midrule
\multirow{4}{*}{1} & Full-comp. timing & 0.3521 & 0.1051 & 0.247 & \underline{0.578} \\
& Uniform-mask & \underline{0.2859} & 0.1613 & 0.274 & 0.548 \\
& ReCache-mask & 0.2992 & \underline{0.1020} & \underline{0.292} & \textbf{0.584} \\
\rowcolor{mainResultCyan}
& \textbf{Joint} & \textbf{0.2779} & \textbf{0.1018} & \textbf{0.300} & \underline{0.581} \\
\bottomrule
\end{tabular}\par
\endgroup
\end{table}

\paragraph{Time-learning comparisons.}
We compare our full method (\emph{Joint}) with three continuous-time controls
on SD3.5 Large under the same generation pipeline and RLOO objective
(\cref{tab:schedule_ablation}). To assess the benefit of adapting both time
families, \emph{Full-computation timing} fixes cached-computation times and
learns full-computation times; crossings can still change the operation order.
Joint achieves lower LPIPS in three settings, while Full-computation timing
yields lower LPIPS at zero-order $K=14$.

\emph{Uniform-mask} and \emph{ReCache-mask} instead optimize the time grid
with Uniform or learned ReCache masks fixed. Both controls share the same $N-1$
continuous degrees of freedom as Joint, whereas only Joint allows the operation
order to change. Joint achieves lower LPIPS than both fixed-mask controls in
all four settings; relative to the two-stage ReCache-mask control, LPIPS
reductions are most pronounced at $K=7$ (5.5\% and 7.1\% for zero- and
first-order caching, versus 0.8\% and 0.2\% at $K=14$). ImageReward also
improves over Uniform-mask in all four settings and over ReCache-mask at $K=7$,
while remaining comparable at $K=14$ ($0.543$ tied and $0.581$ versus $0.584$).

Furthermore, recombining frozen time families without retraining shows that
Joint achieves the lowest mean image cost in seven of eight settings, with both
models exhibiting significant super-additive cost reductions at $K=7$ under
first-order caching after Bonferroni correction (Supplementary Sec.~E).

\Needspace{4\baselineskip}
\begingroup
\widowpenalty=10000
\paragraph{Native-prior comparison.}
At $K=7$, our unshifted
schedules improve ImageReward and PickScore over uniform refreshes on the
native shift-$3$ grid in all four settings (Supplementary Sec.~C).
\par\endgroup

\subsection{Schedule behavior and error allocation}
\label{sec:experiments_analysis}

To relate the final-image objective in \cref{eq:schedule_objective} to
individual updates, we measure local prediction errors and their integration
weights along the learned trajectories.

\paragraph{Evolution of the two point families.}
On SD3 Medium with zero-order caching, training contracts the full-computation
span $L=1-h^{\mathrm c}_K$, forms non-uniform gaps, and moves cached
computations past the final refresh; relative to Uniform, 19 of 21 cached
points use a different preceding full computation. More broadly, the operation
order changes in all eight learned centers (Supplementary Secs.~D--E).

\Needspace{2\baselineskip}
\paragraph{Measuring local prediction error.}
We compare Uniform and Joint on 32 prompts with two seeds each. At each
visited state, we compute a full prediction $v_i^{\mathrm{full}}$ for
diagnosis, keeping both the cache history and trajectory unchanged. Over
cached computations $\mathcal R$, we define local velocity error $r_i$ and
cumulative step-weighted error $A$:
\begin{equation}
\begin{split}
 r_i&=\operatorname{RMS}(\widehat v_i-v_i^{\mathrm{full}}),\\
 A&=\sum_{i\in\mathcal R}h_i r_i,\qquad h_i=|t_{i+1}-t_i|,
\end{split}
\label{eq:local_weighted_error}
\end{equation}
where $h_i r_i$ captures the single-step Euler deviation before rounding.
We also measure the unweighted mean error
$\bar r_R=|\mathcal R|^{-1}\sum_{i\in\mathcal R}r_i$ and the cached
integration span $H_R=\sum_{i\in\mathcal R}h_i$. Across all eight settings,
Joint reduces mean $A$ by 14.7--79.4\% (\cref{tab:error-allocation}), with all
Bonferroni intervals for the change lying strictly below zero (Supplementary
Sec.~E), while the two budgets exhibit distinct error-allocation patterns.

\begin{table}[t]
\centering
\caption{\textbf{Error allocation relative to Uniform.}
Relative changes (\%) in mean cached-computation span $H_R$ and local error
$\bar r_R$, alongside absolute changes in cumulative weighted error $A$ across
the 32-prompt diagnostic set (M/L: Medium/Large; 0/1: caching order; full
results and intervals in Supplementary Sec.~E).}
\label{tab:error-allocation}
\begingroup
\fontsize{9}{10.8}\selectfont
\setlength{\tabcolsep}{3pt}
\renewcommand{\arraystretch}{1.08}
\begin{tabular}{ccrrr}
\toprule
$K$ & \textbf{Model/order}
& $\Delta H_R$ & $\Delta\bar r_R$
& $\Delta A$ \\
\midrule
\multirow{4}{*}{7}
& M/0 & $-27.5$ & $+37.4$ & $-0.0602$ \\
& M/1 & $-51.0$ & $+45.0$ & $-0.2230$ \\
& L/0 & $-92.7$ & $+234.4$ & $-0.2294$ \\
& L/1 & $-66.0$ & $+72.6$ & $-0.1298$ \\
\midrule
\multirow{4}{*}{14}
& M/0 & $+3.7$ & $-1.5$ & $-0.0781$ \\
& M/1 & $-22.6$ & $-40.6$ & $-0.1629$ \\
& L/0 & $-38.1$ & $-7.3$ & $-0.0736$ \\
& L/1 & $-0.5$ & $-7.8$ & $-0.0529$ \\
\bottomrule
\end{tabular}
\endgroup
\end{table}

\paragraph{Reduced cached-computation span at $K=7$.}
At $K=7$, Joint lowers the cumulative step-weighted error $A$ relative to
Uniform even though its mean local error $\bar r_R$ increases. Specifically,
the cached integration span $H_R$ contracts sharply
(\cref{tab:error-allocation}), reducing the total integration weight assigned
to cached predictions.

\paragraph{Local error and interval allocation.}
In contrast, Joint lowers $\bar r_R$ across all four $K=14$ settings. Beyond
mean error and total span, $A$ also depends on how interval lengths pair with
local errors: across both budgets, seven of eight Joint settings systematically
pair larger steps with smaller errors (Supplementary Sec.~E), helping explain
lower $A$ when $\bar r_R$ or $H_R$ changes little. Note that $A$ serves as a
local diagnostic rather than a global error bound; low-precision update errors
and first-order extrapolation diagnostics are reported in Supplementary Sec.~E.

\FloatBarrier
\section{Conclusion}
\label{sec:conclusion}

Our dual-continuous-time formulation jointly learns when to perform full and cached computations under an exact full-computation budget. At comparable latency, it achieves lower LPIPS and higher SSIM across all eight settings, and improves all six fidelity and preference metrics under tighter computation budgets. By combining flexible time parameterization with reinforcement learning from final-image feedback, our method delivers clear visual gains over competing baselines under tight budgets. At the tighter budget, the learned schedules reduce mean cumulative step-weighted local error despite higher mean local error. This observation highlights the role of integration intervals alongside local prediction accuracy in cache scheduling. Future work could extend this perspective to the joint design of caching rules, sampling schedules, and higher-order numerical solvers.

\section*{Acknowledgments}

We used Gemini to check the manuscript's formatting and Codex to help with
\LaTeX{} typesetting, file organization, revision, and consistency checks.
We used OpenAI tools for feedback on drafts and Grammarly for language editing.
Claude and Codex helped us review the experimental scripts and transfer results
into tables. We cross-checked the results using both tools.
We developed the research ideas, hypotheses, claims, and conclusions and
prepared the related work and citations independently of these tools.
We did not use generative AI to generate synthetic data, create datasets, or
annotate data. We manually verified the manuscript and all reported results
and take full responsibility for the work.

{\small
  \bibliographystyle{preprint-fullnames}
  \bibliography{references}
}
\end{document}


\raggedbottom
\maketitle
\appendix
\renewcommand{\thefigure}{S\arabic{figure}}
\renewcommand{\thetable}{S\arabic{table}}
\renewcommand{\thealgorithm}{S\arabic{algorithm}}
Sections~\ref{app:proofs} and~\ref{app:settings} detail the interval policy and experimental settings.
Section~\ref{app:experiments} extends the main results,
\cref{app:figures,app:schedule-analysis} examine training dynamics and learned schedules,
and \cref{app:sd3-qualitative} shows qualitative results.
Section~\ref{app:extended-comparisons} evaluates FLUX.1-dev schedules and two additional SD3 baselines.

Following Sec. 5.3 of the main paper, we refer to our full dual-continuous-time method as Joint throughout the supplementary material (labeled Ours in several comparison tables and qualitative figures).

\section{Interval policy details}
\label{app:proofs}

This section provides the density derivations, policy-gradient formulation,
and schedule-decoding details for our dual-continuous-time parameterization,
where $N$ solver updates include exactly $K$ full computations starting at
$c_0=0$. Full training configurations are detailed in
\cref{app:training_configs}.

\subsection{Policy density}
\label{app:policy-coordinates}

The joint interval action
$h=(h^{\mathrm c},h^{\mathrm r})\in\Delta_+^{K-1}\times\Delta_+^{N-K}$
is mapped to unconstrained coordinates $q=T(h)\in\mathbb R^{N-1}$ drawn from
a diagonal Gaussian $\mathcal N(q;\mu,\Sigma)$. For a $d$-entry simplex
vector, treating $h_{1:d-1}$ as independent variables with
$h_d=1-\sum_{k=1}^{d-1}h_k$, the additive log-ratio (ALR) transform
$a(h)_j=\log h_j-\log h_d$ ($j<d$) has Jacobian entries
$\frac{\partial a_j}{\partial h_k}=\frac{\delta_{jk}}{h_j}+\frac{1}{h_d}$.
By the matrix determinant lemma, its Jacobian determinant is
\begin{equation*}
 \det\!\left(\frac{\partial a(h)}{\partial h_{1:d-1}}\right)
 =\left(\prod_{j=1}^{d-1}\frac{1}{h_j}\right)\!\left(1+\frac{1}{h_d}\sum_{j=1}^{d-1}h_j\right)
 =\prod_{j=1}^{d}\frac{1}{h_j}.
\end{equation*}
For the span--shape encoding (\cref{app:training_configs},
\cref{eq:app_boundary_coordinates}), where $h^{\mathrm c}=(Lw,1-L)$ with
$w\in\Delta_+^{d-2}$ and $q^{\mathrm c}=[\operatorname{logit}L;a(w)]$, the
map from $(L,w_{1:d-2})$ to $h^{\mathrm c}_{1:d-1}$ has absolute determinant
$L^{d-2}$. Combining its reciprocal with
$\frac{\mathrm d\operatorname{logit}L}{\mathrm d L}=\frac{1}{L(1-L)}$ and
$\det(\partial a(w)/\partial w)=\prod_{j=1}^{d-1}w_j^{-1}$ yields
$\frac{1}{1-L}\prod_{j=1}^{d-1}\frac{1}{Lw_j}=\prod_{j=1}^d(h_j^{\mathrm c})^{-1}$.
The start--shape encoding follows analogously by substituting $A$ and $1-A$.
All three coordinate encodings therefore share the same absolute Jacobian
determinant $\prod_{j=1}^d h_j^{-1}$, giving the product-simplex log-density
\begin{equation}
 \log p(h)=\log\mathcal N(q;\mu,\Sigma)
 -\sum_{g\in\{\mathrm c,\mathrm r\}}\sum_j\log h^g_j.
\label{eq:app_policy_density}
\end{equation}
Under the ALR transform, \cref{eq:app_policy_density} recovers the
logistic-normal density~\citep{10.1093/biomet/67.2.261}. Because $h$ and
$q=T(h)$ are detached as fixed samples when evaluating
$\nabla_\phi\log p_\phi(h\mid K)$, the Jacobian term contributes zero
gradient, leaving only the closed-form Gaussian score with respect to
$\mu_\phi(K)$ and any learned exploration scales.

\subsection{Policy optimization}
\label{app:policy-optimization}

\begin{figure*}[!t]
  \centering
  \includegraphics[width=\linewidth]{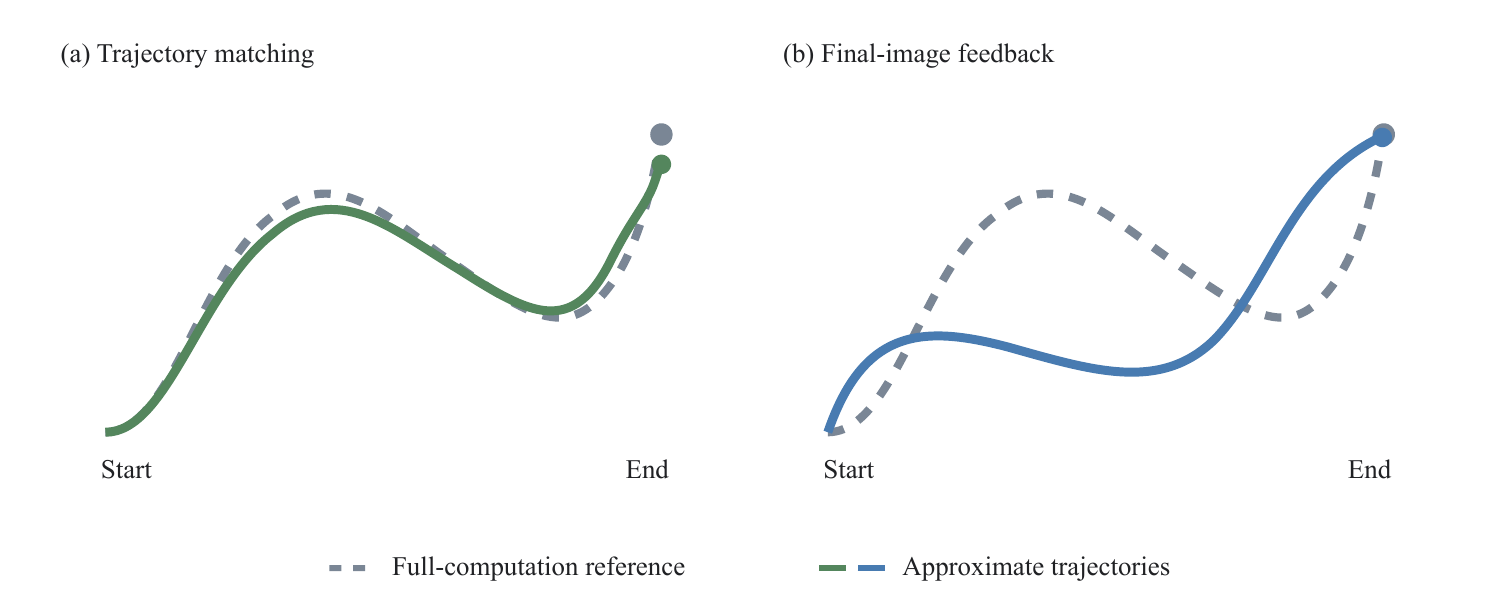}
  \caption{\textbf{Trajectory matching versus final-image feedback.}
  (a) Trajectory matching penalizes deviations from intermediate reference
  states along the denoising path.
  (b) In contrast, our policy optimizes directly for final-image fidelity and
  preference, permitting intermediate trajectory deviations that reach a high-quality endpoint.
  Dashed gray curves denote the full-computation reference, whereas solid
  colored curves denote approximate trajectories from the same initial noise;
  curves and endpoint gaps are schematic.}
  \label{fig:trajectory-objectives}
\end{figure*}

\paragraph{Final-image feedback and sampling.}
Rather than matching intermediate latent trajectories, we optimize both time
families directly from final-image feedback (\cref{fig:trajectory-objectives});
local velocity prediction errors are used solely for post-hoc diagnosis and
do not enter the training loss (\cref{app:analysis-error}). At policy update
$\ell$, candidate schedules are sampled by perturbing the coordinate-space
center $\mu_\phi(K)$ with positive exploration scales $s_\ell$:
\begin{equation}
 q=\mu_\phi(K)+s_\ell\odot\epsilon,\quad
 \epsilon\sim\mathcal N(0,I),\quad h=T^{-1}(q),
\label{eq:app_policy_sampling}
\end{equation}
with initialization and scale-update rules specified in
\cref{app:training_configs}.

\paragraph{Policy-gradient implementation.}
For each condition--noise pair $(c,z)$, the $S$ candidate schedules execute
with independent cache histories while sharing a precomputed full-computation
reference $y^{\mathrm{full}}$. Each candidate incurs a final-image cost
$C_s=D_{\mathrm{patch}}(y_s,y^{\mathrm{full}})-\alpha R_{\mathrm{train}}(y_s,c)$.
Using the leave-one-out baseline $b_s=(S-1)^{-1}\sum_{q\ne s}C_q$ and group
mean $\bar C=S^{-1}\sum_s C_s$, the identity
$C_s-b_s=\frac{S}{S-1}(C_s-\bar C)$ yields the RLOO surrogate loss
\begin{equation}
 \mathcal L_{\mathrm{PG}}
 =\frac{1}{S-1}\sum_{s=1}^{S}
 \operatorname{sg}(C_s-\bar C)\log p_\phi(h_s\mid K),
\label{eq:app_rloo_surrogate}
\end{equation}
where $\operatorname{sg}$ denotes the stop-gradient operator. Evaluated at
fixed sampled actions, gradients update the policy mean and any learned
log-scales without backpropagating through sorting, ODE integration, or
image scoring. Because the simplex parameterization enforces the budget $K$
by construction, no penalty term is needed, and $\mathcal L_{\mathrm{PG}}$
serves as the complete training objective for Joint.

\begin{figure*}[t]
\centering
\makeatletter
\providecommand{\theHALG@line}{}
\renewcommand{\theHALG@line}{\thealgorithm.\arabic{ALG@line}}
\makeatother
\captionsetup[algorithm]{font=normalsize,labelfont=bf,textfont=normalfont,labelsep=space,hypcap=false}
\algrenewcommand\algorithmicindent{1em}
\begin{minipage}[t]{\columnwidth}
  \setlength{\intextsep}{0pt}
  \begin{algorithm}[H]
  \caption{Policy training}
  \label{alg:policy-training}
  \begin{algorithmic}[1]
    \State \textbf{Given:} Training pairs $\mathcal D$ and
      full-computation references $y^{\mathrm{full}}$;
      fixed $N,K$; coordinates $T$; group size $S\geq2$.
    \State \textbf{Return:} Trained policy parameters $\phi$.
    \State Initialize residual means and exploration scales.
    \For{each policy update}
      \State Draw a minibatch $\mathcal B\subset\mathcal D$.
      \For{each $(c,z)\in\mathcal B$}
        \For{$s=1,\ldots,S$}
          \State Sample $h_s=(h_s^{\mathrm c},h_s^{\mathrm r})$
            with independent coordinate noise for the two families
            (\cref{eq:app_policy_sampling}).
          \State $y_s\gets$ Algorithm~\ref{alg:image-sampling} with $(c,z,h_s)$.
          \State Evaluate $C_s=D_{\mathrm{patch}}(y_s,y^{\mathrm{full}})
            -\alpha R_{\mathrm{train}}(y_s,c)$.
        \EndFor
        \State Compute $b_s=\frac{1}{S-1}\sum_{q\ne s}C_q$ for each $s$.
        \State Accumulate the RLOO gradient using $C_s-b_s$
          (\cref{eq:app_rloo_surrogate}).
      \EndFor
      \State Update $\phi$; bound log-scales when learned.
    \EndFor
  \end{algorithmic}
  \end{algorithm}
\end{minipage}\hfill
\begin{minipage}[t]{\columnwidth}
  \setlength{\intextsep}{0pt}
  \begin{algorithm}[H]
  \caption{Image sampling}
  \label{alg:image-sampling}
  \begin{algorithmic}[1]
    \State \textbf{Given:} Policy $\phi$; coordinates $T$; fixed $N,K$;
      condition $c$ and noise $z$; cache backend and map $\psi$.
    \State \textbf{Optional:} Interval action $h$ from
      Algorithm~\ref{alg:policy-training}.
    \State \textbf{Return:} Generated image $y$.
    \State If no interval action is supplied, set
      $h\gets T^{-1}(\mu_\phi(K))$.
    \State Construct $K$ full-computation points (including $c_0=0$)
      and $N-K$ cached-computation points via \cref{eq:app_time_points}.
    \State Sort $\{(c_j,1)\}\cup\{(r_j,0)\}$ by progress to obtain
      $\bigl((u_i,m_i)\bigr)_{i=0}^{N-1}$.
    \State Append $u_N=1$ and map $t_i=\psi(u_i)$.
    \State $x_0\gets z$; initialize empty cache history.
    \For{$i=0,\ldots,N-1$}
      \State Obtain $\widehat v_i$ at $(x_i,t_i,c)$ by a full computation
        if $m_i=1$, otherwise by a cached computation.
      \State Store the full computation's features if $m_i=1$;
        update $x_{i+1}$ via \cref{eq:app_cached_euler}.
    \EndFor
    \State $y\gets\operatorname{Dec}(x_N)$.
  \end{algorithmic}
  \end{algorithm}
\end{minipage}
\end{figure*}

\subsection{Schedule decoding and numerical handling}
\label{app:schedule-decoding}

\paragraph{Schedule construction and deterministic inference.}
Algorithms~\ref{alg:policy-training} and~\ref{alg:image-sampling} share the
same schedule decoder. During inference, we bypass stochastic sampling and
deterministically decode the coordinate-space center
$h^\star=T^{-1}(\mu_\phi(K))$. Given an interval action
$h=(h^{\mathrm c},h^{\mathrm r})$, within-family cumulative sums produce the
two point sets
\begin{equation}
\begin{split}
 c_0&=0,\quad c_j=\sum_{i=1}^j h^{\mathrm c}_i
       \quad (j=1,\ldots,K-1),\\
 r_j&=\sum_{i=1}^j h^{\mathrm r}_i
       \quad (j=1,\ldots,N-K).
\end{split}
\label{eq:app_time_points}
\end{equation}
Merging and sorting $\{(c_j,1)\}\cup\{(r_j,0)\}$ by normalized progress
yields $\bigl((u_i,m_i)\bigr)_{i=0}^{N-1}$, to which we append the terminal
boundary $u_N=1$ and apply $t_i=\psi(u_i)$. Whenever all interior points are
distinct, partitioning by operation label $m_i$ and taking successive
differences uniquely recovers $(h^{\mathrm c},h^{\mathrm r})$. During
sampling, full steps ($m_i=1$) refresh the cache history while cached steps
($m_i=0$) query the caching backend; in either case, the resulting velocity
estimate $\widehat v_i$ drives the Euler update
\begin{equation}
 x_{i+1}=x_i+(t_{i+1}-t_i)\widehat v_i.
\label{eq:app_cached_euler}
\end{equation}

\paragraph{Finite-precision grids.}
Although coincident points occur with probability zero under exact arithmetic,
finite precision can occasionally produce ties. In Joint, FP64 ties on the
progress grid are resolved using adjacent representable values while
preserving endpoints, operation labels, and the exact budget $K$; likewise,
the scheduler map $\psi$ resolves FP32 ties on the decreasing noise-level
grid. Policy log-densities are always evaluated on the raw sampled interval
actions prior to tie-breaking. For the Full-computation timing control,
unordered progress grids are rejected outright so that the fixed
cached-computation points remain unaltered.
\FloatBarrier
\section{Experimental settings}
\label{app:settings}

\subsection{Data, generation, and metrics}

\paragraph{Generation protocol.}
All main comparisons use the SD3 Euler scheduler at $512\times512$ resolution
with $\gamma=1$. Following Stability AI's Diffusers
examples~\citep{stabilityai-sd3-medium-model-card,stabilityai-sd35-large-model-card},
we use $N=28$ steps with classifier-free
guidance~\citep{ho2022classifierfreediffusionguidance}: SD3 Medium runs in
FP16 (guidance scale $7.0$) and SD3.5 Large runs in BF16 (guidance scale
$3.5$), with the stated precision applied to both the generator and initial
noise.

\paragraph{Timestep shifting.}
The native scheduler maps unshifted noise levels $\tau\in[0,1]$
via~\citep{esser2024scaling}
\begin{equation}
 t=\frac{\gamma\tau}{1+(\gamma-1)\tau},\qquad \gamma>0.
 \label{eq:app_timestep_shift}
\end{equation}
When $\gamma=1$, this reduces to the identity map. We define $t=\psi(u)$ by
linearly interpolating the shifted native boundaries from $t=1$ to $t=0$, so
that $u_i=i/N$ recovers the native schedule.

\paragraph{Data and pairing.}
Policy training uses 30,000 fixed caption records from the MS-COCO 2014
validation split~\citep{10.1007/978-3-319-10602-1_48}, retaining duplicate
captions as distinct records. Evaluation pairs 200 DrawBench
prompts~\citep{NEURIPS2022_ec795aea} with five fixed seeds each (1,000 images
per setting). For each training record, all $S$ candidate schedules share the
same prompt, initial latent, and precomputed full-computation reference.

\paragraph{Metrics.}
We evaluate whole-image LPIPS (AlexNet), PSNR, and SSIM on decoded RGB images
against their paired full-computation references. During training,
$D_{\mathrm{patch}}$ averages AlexNet LPIPS over non-overlapping $64\times64$
patches, and $R_{\mathrm{train}}$ uses HPSv2.0 ViT-H/14 (OpenCLIP 3.3.0).

\subsection{Cache and scheduling implementations}

\paragraph{Cache approximation rules.}
Zero-order reuse (FORA-style) copies attention and feed-forward outputs from
the latest full computation across all MMDiT blocks, whereas first-order
forecasting (TaylorSeer-style) linearly extrapolates these outputs from the
two latest full computations and falls back to zero-order reuse when only one
refresh is available. First-order extrapolation uses continuous noise levels
by default; in our SeaCache adapter, first-order forecasting uses discrete
step indices on SD3 Medium at both budgets and on SD3.5 Large at $K=14$.

\paragraph{Scheduling adapters.}
All four main-table baselines operate on the fixed native time grid. Uniform
places full computations at the $K$ reference indices defined in
\cref{app:policy-initialization}, while ReCache~\citep{aliev2026recache}
learns discrete refresh indices offline. For online adaptive methods,
SeaCache~\citep{Chung_2026_CVPR} applies its adapted spectral-evolution
criterion, and DiCache~\citep{bu2026dicache} pairs Online Probe Profiling
(depth $1$, retention $0.2$) with the selected caching backend in place of
Dynamic Cache Trajectory Alignment. We calibrate the online thresholds of both
methods to match a mean budget of $K$ using training records and computation
counts alone; on SD3 Medium, SeaCache's zero-order thresholds are shared
across both backends.

\subsection{Training configurations}
\label{app:training_configs}

\paragraph{Common training settings.}
ReCache and Joint share the same training data, generation settings,
composite image cost ($\alpha=7$), and RLOO policy-gradient estimator, using
eight prompt--noise records per update, $S=6$ candidates per record, and
Adam~\citep{kingma2015iclr-adam} ($\beta=(0.9,0.999)$, $\epsilon=10^{-8}$,
zero weight decay). ReCache additionally applies linearly annealed entropy
regularization and trains for 1,000 updates across all settings, whereas
Joint uses no entropy term and its per-setting update counts (e.g., 1,500 on
SD3 Medium at $K=7$) are listed in \cref{tab:policy_configurations}.

\paragraph{ReCache training.}
We adapt the multi-budget ReCache policy~\citep{aliev2026recache} by training
a dedicated policy per model, backend, and budget $K$. An MLP maps $K$ to
logits over the 28 updates; with step $i=0$ fixed as a full computation, the
remaining $K-1$ full steps are selected via Gumbel top-$(K-1)$ during training
and by the $K-1$ highest logits at inference. Each policy is trained for
1,000 optimizer updates at a learning rate of $5\times10^{-4}$, with an
entropy regularization coefficient decaying linearly from $0.005$ to zero.

\paragraph{Coordinate encodings.}
For a simplex vector $h\in\Delta_+^{d-1}\subset\mathbb R^d$, the additive
log-ratio (ALR) transform~\citep{19b3d08d-3e3e-3ca5-a174-44ff933afefb} and
its inverse are given by
\begin{equation}
 a(h)_j=\log(h_j/h_d),\quad j<d,\qquad
 h=\operatorname{softmax}([a(h);0]),
\label{eq:alr}
\end{equation}
which expresses interval lengths relative to the final gap $h_d$. To decouple
boundary placement from internal spacing, our boundary-split encodings isolate
either the full-computation span $L=1-h^{\mathrm c}_K$ (\emph{span--shape})
or the first cached-computation time $A=h^{\mathrm r}_1$
(\emph{start--shape}). With remainders $w\in\Delta_+^{K-2}$ and
$v\in\Delta_+^{N-K-1}$, these transforms are defined as
\begin{equation}
\begin{aligned}
 h^{\mathrm c}&=(Lw,1-L), & q^{\mathrm c}&=[\operatorname{logit}L;a(w)],\\
 h^{\mathrm r}&=(A,(1-A)v), & q^{\mathrm r}&=[\operatorname{logit}A;a(v)].
\end{aligned}
\label{eq:app_boundary_coordinates}
\end{equation}
Their inverses apply a sigmoid to the boundary coordinate and a softmax to
the shape coordinates, and concatenating the two families yields
$q\in\mathbb R^{N-1}$. Boundary splits require at least two intervals; a
single interval reduces to $(1)$ with zero degrees of freedom, and empty
groups contribute zero noise and zero log-density. Although all three
encodings span the same feasible simplex, boundary splits allow isotropic
Gaussian exploration to adjust the horizon and internal ratios along separate
coordinates.

\begin{table*}[t]
\centering
\caption{\textbf{Policy training configurations for Joint.}
Each model--backend--budget setting is trained with a dedicated policy.
Final training updates indicate the total number of optimizer updates
completed for each setting.}
\label{tab:policy_configurations}
\begingroup
\fontsize{9}{11}\selectfont
\setlength{\tabcolsep}{3pt}
\renewcommand{\arraystretch}{1.1}
\begin{tabular*}{\textwidth}{@{\extracolsep{\fill}}lcccc@{}}
\toprule
\multicolumn{5}{l}{\textbf{(a) SD3 Medium}} \\
\textbf{Parameter} & \textbf{Zero-order, $K=7$} & \textbf{First-order, $K=7$} & \textbf{Zero-order, $K=14$} & \textbf{First-order, $K=14$} \\
\midrule
Full-computation coordinates & Span--shape & Span--shape & ALR & Span--shape \\
Cached-computation coordinates & ALR & ALR & ALR & Start--shape \\
Exploration scales & Learned & Learned & Fixed $0.1$ & Learned \\
Mean learning rate & $5\!\times\!10^{-4}$ & $5\!\times\!10^{-4}$ & $5\!\times\!10^{-4}$ & $5\!\times\!10^{-4}\!\to\!10^{-4}$\textsuperscript{a} \\
Final training update & 1,500 & 1,500 & 600 & 2,400 \\
\midrule
\multicolumn{5}{l}{\textbf{(b) SD3.5 Large}} \\
\textbf{Parameter} & \textbf{Zero-order, $K=7$} & \textbf{First-order, $K=7$} & \textbf{Zero-order, $K=14$} & \textbf{First-order, $K=14$} \\
\midrule
Full-computation coordinates & ALR & Span--shape & Span--shape & ALR \\
Cached-computation coordinates & ALR & ALR & Start--shape & ALR \\
Exploration scales & Fixed $0.1$ & Learned & Learned & Fixed $0.1$ \\
Mean learning rate & $5\!\times\!10^{-4}$ & $10^{-4}$ & $10^{-4}$ & $5\!\times\!10^{-4}\!\to\!5\!\times\!10^{-5}$\textsuperscript{b} \\
Final training update & 1,500 & 912 & 1,500 & 1,000 \\
\bottomrule
\end{tabular*}
\par\smallskip
\begin{minipage}{\textwidth}
\textsuperscript{a}Step-reduced at update 1,800.\quad
\textsuperscript{b}Cosine decay over updates 600--1,000.
\end{minipage}
\endgroup
\end{table*}

\paragraph{Policy and initialization.}
\label{app:policy-initialization}
A lightweight MLP predicts the coordinate residual $f_\phi(K)$ in
$\mu_\phi(K)=T(\bar h)+f_\phi(K)$, initialized to zero so that all internal
times remain trainable around the reference $\bar h$. By default, Joint
initializes from the uniform reference in Sec.~4.3 of the main paper: on the
progress grid $i/N$, full computations occupy indices
$\operatorname{round}(j(N-1)/(K-1))$ for $j=0,\ldots,K-1$ (using
round-to-even, or only step $0$ when $K=1$), with all remaining steps cached.
As the sole exception to the default uniform initialization in Sec.~4.3 of
the main paper, SD3.5 Large with zero-order caching at $K=14$ initializes
with $L=0.82$ and $A=0.28$, spacing full and cached times uniformly over
$[0,L]$ and $[A,1)$, respectively. In all cases, $\bar h$ is extracted from
the resulting within-family intervals and endpoint gaps.

\paragraph{Exploration.}
When exploration scales are learned, we tie them within each boundary or shape
block and update $\eta_g=\log s_g$ alongside the policy means via the RLOO
surrogate (\cref{eq:app_rloo_surrogate}). Scale learning rates and per-step
clips $\delta$ are set to $(0.005,0.02)$ on SD3 Medium and $(0.001,0.01)$ on
SD3.5 Large, with each update clipped to both the global bounds and
$[\eta_g-\delta,\eta_g+\delta]$.

\paragraph{Control settings.}
The Full-computation timing control uses ALR coordinates, a fixed exploration
scale of $0.1$, a constant mean learning rate of $5\times10^{-4}$, and 1,500
updates. Both policies in \cref{fig:dual_schedule_points} come from dedicated
diagnostic runs under matched ALR coordinates.

\paragraph{Recorded training time.}
On a single NVIDIA H200 GPU, first-order policy optimization for Joint took
13.21 hours on SD3 Medium ($K=7$) and 29.96 hours on SD3.5 Large ($K=14$)
under the update counts in \cref{tab:policy_configurations}, excluding
reference generation and post-training evaluation.

\subsection{Configurations for ablations and additional comparisons}

All ablation controls share Joint's generation pipeline, pretrained weights,
$N$, $K$, caching backend, and RLOO objective.

\paragraph{Full-computation timing comparison.}
This control optimizes only $h^{\mathrm c}$ while keeping
$h^{\mathrm r}=\bar h^{\mathrm r}$ fixed at the uniform reference without
exploration or gradient updates; full-computation points can still cross the
fixed cached-computation points and alter the operation labels $m_i$.

\paragraph{Fixed-mask time-learning controls.}
\label{app:fixed_mask_control}
Uniform-mask and ReCache-mask optimize a single ordered time grid under a
fixed operation mask $\bar m\in\{0,1\}^N$. Unconstrained ALR coordinates
$q\in\mathbb R^{N-1}$ map to interval lengths
$g=\operatorname{softmax}([q;0])$ and progress boundaries
$u_i=\sum_{j=1}^{i}g_j$ for $i=1,\ldots,N$ (with terminal boundary $u_N=1$),
executing operation $m_i=\bar m_i$ at each step. Both controls therefore match
Joint's $N-1$ continuous degrees of freedom and exact budget $K$.

Uniform-mask adopts the uniform reference labels from
\cref{app:policy-initialization}, whereas ReCache-mask first learns $\bar m$
for 1,000 updates (\cref{app:training_configs}) and then optimizes the time
grid for 500 updates with $\bar m$ fixed (1,500 updates total). Table~3 of the
main paper reports LPIPS and ImageReward across 1,000 SD3.5 Large images per
setting for both backends and budgets.

\paragraph{Native-prior comparison.}
At $N=28$ and $K=7$, we compare uniform refreshes on the native shifted grid
($\gamma=3$) against Joint learned and evaluated at $\gamma=1$ using the same
caching backends and 1,000 DrawBench prompt--noise pairs. Because different
timestep shifts produce different full-computation references, we evaluate
this comparison with reference-free ImageReward and PickScore.
\FloatBarrier
\section{Additional experimental results}
\label{app:experiments}

All additional comparisons below follow the evaluation protocol and training
schedules detailed in \cref{app:settings,tab:policy_configurations}.

\subsection{Image cost and quality trade-offs}
\label{app:additional-metrics}

\ifdefined\table
  \let\endAdditionalMetricsPreview\relax
\else
  \documentclass[10pt,letterpaper]{article}
  \usepackage[textwidth=6.875in,vmargin=0.75in]{geometry}
  \usepackage{booktabs,times,xcolor,colortbl,multirow}
  \pagestyle{empty}
  \def\endAdditionalMetricsPreview{\end{document}}
  \begin{document}
\fi
\begin{table*}[t]
\centering
\caption{\textbf{Composite image cost and realized full-computation counts.}
We report the mean composite cost $C=D_{\mathrm{patch}}-7R_{\mathrm{train}}$
($\downarrow$) alongside the realized mean full-computation counts of online
adaptive policies across the 1,000-image DrawBench benchmark.}
\label{tab:schedule_additional_metrics}
\begingroup
\fontsize{9}{10.8}\selectfont
\setlength{\tabcolsep}{4pt}
\renewcommand{\arraystretch}{1.10}
\begin{tabular*}{\textwidth}{@{\extracolsep{\fill}}llrrrrrrr@{}}
\toprule
& & & \multicolumn{4}{c}{\textbf{Image cost $C\downarrow$}}
    & \multicolumn{2}{c}{\textbf{Mean full computations}} \\
\cmidrule(lr){4-7}\cmidrule(lr){8-9}
\textbf{Model} & \textbf{Cache backend} & \textbf{\boldmath$K$}
& \textbf{SeaCache} & \textbf{ReCache} & \textbf{Joint}
& \textbf{Full-comp. timing} & \textbf{SeaCache} & \textbf{DiCache} \\
\midrule
\multirow{4}{*}{SD3 Medium} & Zero-order & 7 & -1.5540 & -1.7376 & -1.7565 & --- & 6.981 & 7.000 \\
 & Zero-order & 14 & -1.8722 & -1.8808 & -1.8873 & --- & 14.000 & 13.922 \\
 & First-order & 7 & -1.4383\rlap{$^{\dagger}$} & -1.5895 & -1.7812 & --- & 6.991 & 7.000 \\
 & First-order & 14 & -1.8907\rlap{$^{\dagger}$} & -1.8882 & -1.8928 & --- & 13.997 & 13.951 \\
\midrule
\multirow{4}{*}{SD3.5 Large} & Zero-order & 7 & -1.6183 & -1.7208 & -1.7219 & -1.6982 & 7.012 & 7.000 \\
 & Zero-order & 14 & -1.8643 & -1.8792 & -1.8809 & -1.8842 & 14.000 & 14.224 \\
 & First-order & 7 & -1.5195 & -1.6953 & -1.7381 & -1.6722 & 6.935 & 7.002 \\
 & First-order & 14 & -1.9014\rlap{$^{\dagger}$} & -1.9019 & -1.9036 & -1.8974 & 14.000 & 14.090 \\
\bottomrule
\end{tabular*}
\par\smallskip
\begin{minipage}{\linewidth}
\fontsize{8}{9.4}\selectfont
Static policies (ReCache, Joint, and Full-computation timing) execute exactly
$K$ full computations per trajectory; Full-computation timing is evaluated on
SD3.5 Large only.\par
$^{\dagger}$ SeaCache uses discrete step indices for first-order extrapolation
in the marked settings, whereas all other first-order runs use continuous noise
levels.
\end{minipage}
\endgroup
\end{table*}
\endAdditionalMetricsPreview

\paragraph{Composite image cost.}
Across all eight settings in the main tables, all five policies are evaluated
on the same 1,000 DrawBench images. Relative to per-budget ReCache, Joint
achieves a lower mean composite cost $C=D_{\mathrm{patch}}-7R_{\mathrm{train}}$
in every setting (\cref{tab:schedule_additional_metrics}). Note that rankings
under $C$ can differ slightly from those under whole-image LPIPS, as $C$
jointly balances patchwise perceptual fidelity and human preference.

\paragraph{Fidelity and preference trade-offs at $K=14$.}
Under the higher budget of $K=14$ (50\% full computations), Joint consistently
improves LPIPS, SSIM, and composite cost $C$ over ReCache across all four
model--backend settings, while PSNR increases on SD3 Medium and dips slightly
on SD3.5 Large. Preference metrics exhibit minor, metric-specific trade-offs:
on SD3 Medium, zero-order caching improves ImageReward at a slight cost to
PickScore, whereas first-order caching shows the reverse pattern; on SD3.5
Large, zero-order caching improves both metrics, while first-order caching
matches PickScore at reported precision with a slight drop in ImageReward.
HPSv2 is marginally lower across all four settings.

\subsection{\texorpdfstring{Full-computation budgets and\\end-to-end latency}{Full-computation budgets and end-to-end latency}}
\label{app:compute-latency}

\ifdefined\table
  \let\endLatencyPreview\relax
\else
  \documentclass[10pt,letterpaper]{article}
  \usepackage[textwidth=6.875in,vmargin=0.75in]{geometry}
  \usepackage{booktabs,times,xcolor,colortbl,multirow}
  \pagestyle{empty}
  \def\endLatencyPreview{\end{document}}
  \begin{document}
\fi
\definecolor{mainResultCyan}{rgb}{0.925,1,1}

\begin{table*}[t]
\centering
\caption{\textbf{End-to-end inference latency and speedup.}
We report mean latency in seconds ($\downarrow$) averaged over 100 runs, with
speedups ($\uparrow$) relative to full 28-step inference shown in parentheses.
Cached schedules use $N=28$ with target budgets $K\in\{7,14\}$, whereas the
uncached reference uses $N=K=28$.}
\label{tab:latency}
\begingroup
\fontsize{9}{10.8}\selectfont
\setlength{\tabcolsep}{3.2pt}
\renewcommand{\arraystretch}{1.12}
\begin{tabular}{@{}>{\centering\arraybackslash}p{0.24\textwidth}*{4}{>{\centering\arraybackslash}p{\dimexpr0.19\textwidth-2\tabcolsep\relax}}@{}}
\toprule
\multirow{2}{*}{\textbf{Sampling policy}}
& \multicolumn{2}{c}{\textbf{SD3 Medium}}
& \multicolumn{2}{c}{\textbf{SD3.5 Large}} \\
\cmidrule(lr){2-3}\cmidrule(lr){4-5}
& \textbf{\boldmath$K{=}7$} & \textbf{\boldmath$K{=}14$}
& \textbf{\boldmath$K{=}7$} & \textbf{\boldmath$K{=}14$} \\
\midrule
\multicolumn{5}{@{}l@{}}{\textbf{Uncached references}} \\
Full inference
& \multicolumn{2}{c}{0.779 (1.00$\times$)}
& \multicolumn{2}{c}{2.395 (1.00$\times$)} \\
\midrule
\multicolumn{5}{@{}l@{}}{\textbf{(a) Zero-order caching}} \\
Uniform
& 0.365 (2.13$\times$) & 0.506 (1.54$\times$)
& 0.906 (2.64$\times$) & 1.398 (1.71$\times$) \\
DiCache
& 0.392 (1.99$\times$) & 0.527 (1.48$\times$)
& 0.963 (2.49$\times$) & 1.437 (1.67$\times$) \\
SeaCache
& 0.380 (2.05$\times$) & 0.523 (1.49$\times$)
& 0.925 (2.59$\times$) & 1.420 (1.69$\times$) \\
ReCache
& 0.364 (2.14$\times$) & 0.506 (1.54$\times$)
& 0.912 (2.63$\times$) & 1.405 (1.70$\times$) \\
\rowcolor{mainResultCyan}
\textbf{Ours}
& 0.365 (2.14$\times$) & 0.507 (1.54$\times$)
& 0.913 (2.62$\times$) & 1.401 (1.71$\times$) \\
\midrule
\multicolumn{5}{@{}l@{}}{\textbf{(b) First-order caching}} \\
Uniform
& 0.469 (1.66$\times$) & 0.580 (1.34$\times$)
& 1.092 (2.19$\times$) & 1.529 (1.57$\times$) \\
DiCache
& 0.509 (1.53$\times$) & 0.606 (1.29$\times$)
& 1.181 (2.03$\times$) & 1.575 (1.52$\times$) \\
SeaCache
& 0.437 (1.78$\times$) & 0.574 (1.36$\times$)
& 1.141 (2.10$\times$) & 1.503 (1.59$\times$) \\
ReCache
& 0.484 (1.61$\times$) & 0.584 (1.33$\times$)
& 1.129 (2.12$\times$) & 1.546 (1.55$\times$) \\
\rowcolor{mainResultCyan}
\textbf{Ours}
& 0.473 (1.65$\times$) & 0.584 (1.33$\times$)
& 1.090 (2.20$\times$) & 1.535 (1.56$\times$) \\
\bottomrule
\end{tabular}
\par\smallskip
\begin{minipage}{\textwidth}
\fontsize{8}{9.6}\selectfont
Measured on a single NVIDIA RTX Pro 6000 GPU with batch size 1. End-to-end
timing includes prompt encoding, denoising, VAE decoding, as well as the
online scheduling decisions and probes of SeaCache and DiCache.
\end{minipage}
\endgroup
\end{table*}
\endLatencyPreview

\paragraph{Computation accounting.}
Full-computation counts include the initial step $t_0$ and count both
classifier-free guidance branches together as a single evaluation, while the
terminal boundary $u_N=1$ only closes integration. Static policies (Uniform,
ReCache, Joint, Full-computation timing, Uniform-mask, and ReCache-mask)
execute exactly $K$ full and $28-K$ cached computations per trajectory. In
contrast, SeaCache and DiCache target a mean budget of $K$ via calibrated
online thresholds, with their realized means over 1,000 images reported in
\cref{tab:schedule_additional_metrics}; DiCache additionally runs one probe
per cached step (averaging $28$ minus the mean full count). End-to-end
latencies in \cref{tab:latency} are averaged over 100 runs per setting and
include all online decisions, probes, and cached updates.

\paragraph{Major-operator computation.}
On the fixed input profiled in \cref{tab:flops}, Joint reduces counted
end-to-end FLOPs by 46.9--73.6\% relative to full 28-step inference, matching
Uniform and ReCache at each model and budget. By comparison, DiCache's online
probes add 2.6--9.9\% FLOPs overhead over Joint.

\begin{table*}[t]
\centering
\caption{\textbf{Major-operator FLOPs of cache-scheduling methods.}
All methods are profiled over $N=28$ updates at $512\times512$ resolution
with batch size 1 on a fixed prompt and initial noise. FLOPs are reported in
TFLOPs/image ($10^{12}$ operations), and compression denotes the reduction
factor relative to full 28-step inference on the same model. Full and Probe
indicate the executed full-network and additional probe-block counts,
respectively.}
\label{tab:flops}
\begingroup
\fontsize{9}{10.8}\selectfont
\setlength{\tabcolsep}{3pt}
\renewcommand{\arraystretch}{1.12}
\begin{tabular*}{\textwidth}{@{\extracolsep{\fill}}lccrrrrrr@{}}
\toprule
& \multicolumn{2}{c}{\textbf{Executions}}
& \multicolumn{3}{c}{\textbf{SD3 Medium}}
& \multicolumn{3}{c}{\textbf{SD3.5 Large}} \\
\cmidrule(lr){2-3}\cmidrule(lr){4-6}\cmidrule(lr){7-9}
\textbf{Policy} & \textbf{Full} & \textbf{Probe}
& \textbf{Denoising} & \textbf{End-to-end} & \textbf{Compression}
& \textbf{Denoising} & \textbf{End-to-end} & \textbf{Compression} \\
\midrule
Full inference & 28 & 0
& 118.017 & 125.548 & 1.00$\times$
& 446.756 & 454.287 & 1.00$\times$ \\
\midrule
\multicolumn{9}{@{}l}{\textbf{Caching at $K=7$}} \\
Uniform, ReCache, Joint & 7 & 0
& 29.755 & 37.285 & 3.37$\times$
& 112.221 & 119.752 & 3.79$\times$ \\
DiCache & 7 & 21
& 33.459 & 40.990 & 3.06$\times$
& 121.070 & 128.601 & 3.53$\times$ \\
SeaCache & 7 & 0
& 29.756 & 37.287 & 3.37$\times$
& 112.225 & 119.756 & 3.79$\times$ \\
\midrule
\multicolumn{9}{@{}l}{\textbf{Caching at $K=14$}} \\
Uniform, ReCache, Joint & 14 & 0
& 59.175 & 66.706 & 1.88$\times$
& 223.733 & 231.263 & 1.96$\times$ \\
DiCache & 14 & 14
& 61.645 & 69.176 & 1.81$\times$
& 229.632 & 237.163 & 1.92$\times$ \\
SeaCache & 14 & 0
& 59.177 & 66.708 & 1.88$\times$
& 223.737 & 231.267 & 1.96$\times$ \\
\bottomrule
\end{tabular*}
\par\smallskip
\begin{minipage}{\textwidth}
\fontsize{8}{9.6}\selectfont
End-to-end FLOPs cover prompt encoding, denoising, and VAE decoding, counting
each multiply--add as two operations. Included operators are GEMM,
convolution, and attention QK/AV, whereas normalization, softmax, elementwise
operations, FFT, data movement, and CPU control are excluded. Counts are
identical across three profiling repeats and between zero- and first-order
backends, so they are reported once. Dynamic full/probe counts reflect this
fixed input rather than dataset averages. All methods use the scheduling
adapters evaluated in \cref{tab:latency}, which reports the corresponding
measured latency and speedup.
\end{minipage}
\endgroup
\end{table*}

\paragraph{Latency comparisons.}
Because Joint precomputes its schedule offline with zero online decision
overhead, it achieves strong wall-clock efficiency (\cref{tab:latency}). For
instance, on SD3.5 Large with first-order caching at $K=7$, Joint reduces
mean end-to-end latency by $3.45\%$, $4.48\%$, and $7.68\%$ relative to
ReCache, SeaCache, and DiCache, respectively, while simultaneously
outperforming ReCache on all six fidelity and preference metrics.

\subsection{Learned timing versus the native prior}
\label{app:native-prior}

Under the paired protocol in Sec.~B.4 ($N=28$, $K=7$), Joint learned and
evaluated on the unshifted grid ($\gamma=1$) outperforms uniform refreshes on
the native shifted grid ($\gamma=3$) in both ImageReward and PickScore across
all four model--backend settings (\cref{tab:schedule_prior}).

\definecolor{mainResultCyan}{rgb}{0.925,1,1}
\begin{table}[H]
\centering
\caption{\textbf{Learned timing versus the native prior.}
Comparison on 1,000 shared DrawBench prompt--noise pairs at $N=28$ and $K=7$.
Default applies uniform refreshes on the native shifted grid ($\gamma=3$),
whereas Ours (Joint) uses schedules learned and evaluated on the unshifted
grid ($\gamma=1$). Bold marks the higher score per model and backend.}
\label{tab:schedule_prior}
\begingroup
\fontsize{9}{10.8}\selectfont
\setlength{\tabcolsep}{3pt}
\renewcommand{\arraystretch}{1.12}
\textbf{(a) Zero-order}\par
\vspace{3pt}
\begin{tabular}{*{4}{c}}
\toprule
\textbf{Model} & \textbf{Policy}
& \textbf{ImageReward\boldmath$\uparrow$} & \textbf{PickScore\boldmath$\uparrow$} \\
\midrule
& Default & -0.020 & 20.70 \\
\rowcolor{mainResultCyan}
\cellcolor{white}\multirow{-2}{*}{SD3 Medium} & \textbf{Ours} & \textbf{0.298} & \textbf{21.17} \\
\cmidrule(lr){2-4}
& Default & 0.260 & 21.26 \\
\rowcolor{mainResultCyan}
\cellcolor{white}\multirow{-2}{*}{SD3.5 Large} & \textbf{Ours} & \textbf{0.323} & \textbf{21.47} \\
\bottomrule
\end{tabular}\par
\vspace{10pt}
\textbf{(b) First-order}\par
\vspace{3pt}
\begin{tabular}{*{4}{c}}
\toprule
\textbf{Model} & \textbf{Policy}
& \textbf{ImageReward\boldmath$\uparrow$} & \textbf{PickScore\boldmath$\uparrow$} \\
\midrule
& Default & -0.472 & 20.13 \\
\rowcolor{mainResultCyan}
\cellcolor{white}\multirow{-2}{*}{SD3 Medium} & \textbf{Ours} & \textbf{0.350} & \textbf{21.18} \\
\cmidrule(lr){2-4}
& Default & 0.136 & 20.97 \\
\rowcolor{mainResultCyan}
\cellcolor{white}\multirow{-2}{*}{SD3.5 Large} & \textbf{Ours} & \textbf{0.300} & \textbf{21.24} \\
\bottomrule
\end{tabular}\par
\endgroup
\end{table}

\setlength{\dbltextfloatsep}{12pt plus 2pt minus 2pt}
\setlength{\dblfloatsep}{8pt plus 2pt minus 2pt}
\section{Training dynamics and schedule evolution}
\label{app:figures}

\begin{figure*}[!tb]
  \input{figures/training_cost}
\end{figure*}

\paragraph{Optimization dynamics.}
Across all eight model--backend--budget settings (\cref{fig:training-cost}),
the mean sampled-schedule cost $C$ exhibits a consistent optimization
trajectory, dropping rapidly during early updates and continuing to decrease
more gradually later in training.

\begin{figure*}[!tb]
  \begin{minipage}[t]{\columnwidth}
    \input{figures/dual_schedule_points}
  \end{minipage}\hfill
  \begin{minipage}[t]{\columnwidth}
    \input{figures/schedule_evolution}
  \end{minipage}
\end{figure*}

\paragraph{Structure of the learned schedules.}
To compare Full-computation timing and Joint in a coordinate-matched
diagnostic setting, \cref{fig:dual_schedule_points} examines dedicated runs
trained under matched ALR coordinates (\cref{app:training_configs}). Matching
points by their within-family order and measuring displacement in units of
$1/28$ in normalized progress $u$, switching from Full-computation timing to
Joint shifts the full-computation points by a mean absolute displacement of
$0.24$ grid units under zero-order caching and $0.80$ grid units under
first-order caching. This substantially larger shift for first-order caching
is consistent with feature extrapolation depending directly on the relative
spacing between refreshes and cached computations.

\paragraph{Schedule evolution.}
\Cref{fig:schedule-evolution} illustrates how the deterministic center schedule
evolves during training on SD3 Medium with zero-order caching ($K=7$).
Adaptation proceeds in two phases: the full-computation span contracts
sharply toward earlier progress within the first 300--600 updates, after which
the cached-computation points and their interleaving with full computations
continue to refine through update 1,500.
\section{Diagnostics of learned schedules}
\label{app:schedule-analysis}

This section analyzes the learned schedules in detail, examining local error
weighting (\cref{app:analysis-error}), operation-order changes
(\cref{app:analysis-order}), interactions between the two time families
(\cref{app:analysis-interaction}), and first-order history geometry
(\cref{app:analysis-history}).

\subsection{Shared protocol}
\label{app:analysis-protocol}

We freeze Uniform ($00$) and the reported Joint centers ($11$) across both
SD3 models, caching orders, and budgets $K\in\{7,14\}$ at $N=28$. All
diagnostic runs use $\gamma=1$, the precision and guidance settings in
\cref{app:settings}, and an exact budget of $K$ full and $28-K$ cached
computations on paired inputs.

\Cref{app:analysis-interaction} evaluates 100 sampled DrawBench prompt IDs
(out of 200, seed 20261020) with two fixed initial noises per prompt, whereas
\cref{app:analysis-error,app:analysis-history} use a 32-prompt subset with
the same noises. Operation-order statistics are computed directly on the
center schedules and verified across all recorded trajectories
(\cref{app:analysis-order}).

We evaluate the composite cost $C=D_{\mathrm{patch}}-7R_{\mathrm{train}}$
and average the two noises within each prompt prior to paired prompt-level
bootstrapping (100,000 draws, seed 20261020). Bonferroni-corrected intervals
cover three separate test families: eight local-error contrasts (99.375\%),
12 interaction and approximation-order contrasts (99.5833\%), and 32
history-geometry correlations (99.84375\%). Center-schedule counts are exact,
and sampling probabilities report Wilson Monte Carlo intervals.

\subsection{Local error and integration weights}
\label{app:analysis-error}

\paragraph{Same-state measurements.}
While Uniform and Joint follow their own time grids, we diagnose local
approximation error by querying $v_i^{\mathrm{full}}$ at each visited state
$x_i$, time $t_i$, and condition $c$. The trajectory update still uses the
original $\widehat v_i$ and leaves the cache history untouched; across all
1,024 trajectories (28,672 measured steps), the resulting images match the
main evaluation pixel for pixel.

\Needspace{9\baselineskip}
Over the set of cached-computation steps $\mathcal R$, we compute
\begin{equation}
\begin{split}
 r_i&=\operatorname{RMS}(\widehat v_i-v_i^{\mathrm{full}}),\quad
 h_i=|t_{i+1}-t_i|,\\
 A&=\sum_{i\in\mathcal R}h_i r_i,\qquad
 H_R=\sum_{i\in\mathcal R}h_i,\\
 \bar r_R&=\frac{1}{|\mathcal R|}\sum_{i\in\mathcal R}r_i,\qquad
 Q=\frac{A}{H_R\bar r_R}.
\end{split}
\label{eq:app_local_error}
\end{equation}
Here, $\operatorname{RMS}$ is taken over all latent channels and spatial
locations of the guided velocity, and $H_R$ measures the total noise-level
span of cached updates. Because holding $h_i$ fixed and uniformly permuting
$r_i$ across cached steps yields an expected weighted error of
$H_R\bar r_R$, a ratio $Q<1$ indicates a negative association between step
size and local error along that trajectory. We compute $Q$ per trajectory
before averaging.

\begin{table*}[!tp]
\centering
\caption{\textbf{Local error and integration weights.}
Comparison of Uniform ($00$) and Joint ($11$) across 32 DrawBench prompts
with two initial noises per prompt (order 0/1 denotes zero-/first-order
caching). Paired confidence intervals for $\Delta A = A_{11} - A_{00}$ apply
the eight-comparison Bonferroni correction (99.375\%). We also report the
mean error--weight ratio $Q_{11}$ for Joint alongside changes in actual
update error $\sum U_i$ and composite image cost $C$ relative to Uniform;
corresponding point estimates for $\Delta H_R$, $\Delta\bar r_R$, and
$\Delta A$ appear in Table~4 of the main paper.}
\label{tab:analysis-local-error}
\begingroup
\fontsize{9}{10.8}\selectfont
\setlength{\tabcolsep}{5pt}
\renewcommand{\arraystretch}{1.06}
\begin{tabular}{lrrrrrr}
\toprule
& \multicolumn{3}{c}{Step-weighted error} & \multicolumn{3}{c}{Additional diagnostics} \\
\cmidrule(lr){2-4}\cmidrule(lr){5-7}
Model / order / $K$ & $A_{00}$ & $A_{11}$ & 99.375\% CI for $\Delta A$
& $Q_{11}$ & $\Delta\sum U_i$ (\%) & $\Delta C$ \\
\midrule
Medium / 0 / 7 & 0.4097 & 0.3496 & $[-0.0975, -0.0187]$ & 0.855 & $-14.7$ & $-0.5484$ \\
Medium / 0 / 14 & 0.2325 & 0.1544 & $[-0.0894, -0.0660]$ & 0.652 & $-33.6$ & $-0.3684$ \\
Medium / 1 / 7 & 0.4887 & 0.2658 & $[-0.2590, -0.1881]$ & 0.763 & $-45.5$ & $-0.5124$ \\
Medium / 1 / 14 & 0.2807 & 0.1178 & $[-0.1811, -0.1444]$ & 0.911 & $-57.9$ & $-0.4100$ \\
Large / 0 / 7 & 0.2891 & 0.0597 & $[-0.2438, -0.2160]$ & 0.838 & $-78.5$ & $-0.3664$ \\
Large / 0 / 14 & 0.1372 & 0.0636 & $[-0.0788, -0.0680]$ & 0.809 & $-50.4$ & $-0.3064$ \\
Large / 1 / 7 & 0.3236 & 0.1939 & $[-0.1489, -0.1111]$ & 1.020 & $-38.5$ & $-0.3395$ \\
Large / 1 / 14 & 0.1595 & 0.1067 & $[-0.0592, -0.0456]$ & 0.730 & $-32.2$ & $-0.3022$ \\
\bottomrule
\end{tabular}
\endgroup
\end{table*}

\paragraph{Observed allocation patterns.}
\Cref{tab:analysis-local-error} extends Table~4 of the main paper with
complete metrics and Bonferroni intervals. At $K=7$, Joint reduces the
cached integration span $H_R$ from $0.777$ under Uniform to
$0.0571$--$0.563$ across the four settings, with lower $A$ despite higher
mean local error $\bar r_R$.

Mean $Q$ falls below one in seven of the eight settings. The sole exception is
SD3.5 Large with first-order caching at $K=7$, where mean $Q$ is $1.020$.

\paragraph{Actual updates and image costs.}
We further define $U_i$ as the RMS difference between the next latents
produced by the actual low-precision Euler updates using $\widehat v_i$
versus $v_i^{\mathrm{full}}$ (full steps yield identical predictions). Across
all eight settings on the 32-prompt subset, Joint reduces
$\sum_{i\in\mathcal R}U_i$ by 14.7--78.5\% while simultaneously lowering $C$
(\cref{tab:analysis-local-error}). Although $A$ is a local diagnostic that
omits error cancellation and propagation, $A$, $\sum_{i\in\mathcal R}U_i$,
and $C$ decrease concurrently across all eight settings.

\subsection{Operation order and cache ownership}
\label{app:analysis-order}

Matching points across schedules by their within-family order, let
$a_k\in\{0,\ldots,K-1\}$ index the latest full computation preceding cached
point $r_k$ ($k=1,\ldots,M$, with $M=N-K$ and fixed $c_0=0$). Between Joint
($11$) and Uniform ($00$), we measure the changed-owner fraction and
normalized cross-family precedence distance:
\begin{equation}
\begin{split}
 F_{\mathrm{owner}}&=\frac1M\sum_{k=1}^{M}
  \mathbf 1\{a_k^{11}\ne a_k^{00}\},\\
 d_{\mathrm{CR}}&=\frac{\sum_{k=1}^{M}|a_k^{11}-a_k^{00}|}{(K-1)M}.
\end{split}
\label{eq:app_ownership}
\end{equation}
The numerators count reassigned cached points and crossed non-initial
refresh boundaries, respectively. As noted in Sec.~B.2, first-order
forecasting draws from refreshes $a_k$ and $a_k-1$ whenever at least two
prior refreshes exist, falling back to zero-order reuse otherwise.

\begin{table*}[!tp]
\centering
\caption{\textbf{Operation order and cache ownership.}
(a) Comparison of final Joint center schedules against Uniform (order
0/1 denotes zero-/first-order caching), reporting reassigned cache owners
($F_{\mathrm{owner}}$) and crossed non-initial full--cached pairs
($d_{\mathrm{CR}}$) as fractions of $M$ cached steps and $(K-1)M$ pairs,
respectively.
(b) Comparison of 100,000 sampled schedules against each final policy
center, reporting 95\% Wilson Monte Carlo confidence intervals for the
probability of at least one order change.}
\label{tab:analysis-order}
\begingroup
\fontsize{9}{10.8}\selectfont
\setlength{\tabcolsep}{3.5pt}
\renewcommand{\arraystretch}{1.06}
\begin{minipage}[t]{.38\textwidth}
\centering
\textbf{(a) Final centers}\par\vspace{3pt}
\begin{tabular}{lrr}
\toprule
Model / order / $K$ & \shortstack[r]{Changed\\ owners} & \shortstack[r]{Changed\\ pairs} \\
\midrule
Medium / 0 / 7 & 19/21 & 28/126 \\
Medium / 0 / 14 & 12/14 & 19/182 \\
Medium / 1 / 7 & 19/21 & 57/126 \\
Medium / 1 / 14 & 14/14 & 66/182 \\
Large / 0 / 7 & 14/21 & 25/126 \\
Large / 0 / 14 & 14/14 & 70/182 \\
Large / 1 / 7 & 18/21 & 41/126 \\
Large / 1 / 14 & 11/14 & 18/182 \\
\bottomrule
\end{tabular}
\end{minipage}\hfill
\begin{minipage}[t]{.60\textwidth}
\centering
\textbf{(b) Samples from final policies}\par\vspace{3pt}
\begin{tabular}{lrrrr}
\toprule
Model / order / $K$ & \shortstack[r]{Any order\\ change} & \shortstack[r]{95\% MC CI\\ (\%)} & \shortstack[r]{Mean changed\\ owners} & \shortstack[r]{Distinct\\ masks} \\
\midrule
Medium / 0 / 7 & 64.70\% & $[64.40, 65.00]$ & 1.228 & 109 \\
Medium / 0 / 14 & 91.60\% & $[91.42, 91.77]$ & 2.376 & 629 \\
Medium / 1 / 7 & 71.87\% & $[71.59, 72.15]$ & 1.066 & 30 \\
Large / 1 / 14 & 86.67\% & $[86.45, 86.87]$ & 1.874 & 426 \\
\bottomrule
\end{tabular}
\end{minipage}
\endgroup
\end{table*}

Relative to Uniform, all eight Joint center schedules alter the operation
order, reassigning 66.7--100\% of cached steps (\cref{tab:analysis-order}a),
with exact agreement across all 1,024 recorded trajectories. Moreover,
sampling 100,000 schedules (seed 20261020) from each of four final policies
produces at least one transposition relative to the policy center in
64.7--91.6\% of draws, with 1.07--2.38 changed owners on average
(\cref{tab:analysis-order}b), reflecting active local crossings. Operation
order also evolves across saved training checkpoints (recorded more densely
than the six snapshots in \cref{fig:schedule-evolution}), exhibiting 15
adjacent-checkpoint order transitions for SD3 Medium zero-order $K=7$, three
for SD3 Medium zero-order $K=14$, and ten for SD3.5 Large first-order $K=14$.

\subsection{Cross-combination of frozen time families}
\label{app:analysis-interaction}

To evaluate cross-family synergy without retraining, we merge and sort each
pair $(\mathcal C_i,\mathcal R_j)$ from the frozen Uniform and Joint time
families in a $2\times2$ factorial design (\cref{tab:analysis-family}). The
four arms generate 6,400 cached images sharing 400 full-computation
references, complementing the retrained controls in Table~3 of the main paper
(Sec.~B.4).

We define the within-order interaction $I$ and between-order difference $D$ as
\begin{equation}
 I=C_{11}-C_{10}-C_{01}+C_{00},\qquad D=I_{\mathrm{first}}-I_{\mathrm{zero}}.
\label{eq:app_interaction}
\end{equation}
A negative interaction ($I<0$) indicates a super-additive benefit, where the
signed joint cost change $(C_{11}-C_{00})$ is more negative than the sum of
the individual signed changes $(C_{10}-C_{00})$ and $(C_{01}-C_{00})$.

\begin{table*}[!tp]
\centering
\caption{\textbf{Cross-combination of frozen time families.}
Mean composite image costs $C$ ($\downarrow$) across 100 DrawBench prompts
with two initial noises per prompt (M/L: SD3 Medium/SD3.5 Large; 0/1:
zero-/first-order caching). Arms $00$ and $11$ denote Uniform and Joint,
respectively, while mixed arms $10$ and $01$ replace Uniform's full- or
cached-computation times with Joint's at the same budget $K$. Paired
differences report unadjusted 95\% confidence intervals;
\cref{fig:analysis-family-interaction} applies the separate 12-comparison
Bonferroni correction to the interaction contrasts.}
\label{tab:analysis-family}
\begingroup
\fontsize{9}{10.8}\selectfont
\setlength{\tabcolsep}{3.5pt}
\renewcommand{\arraystretch}{1.06}
\begin{tabular}{lrrrr@{\hspace{12pt}}rr}
\toprule
& \multicolumn{4}{c}{Image costs} & \multicolumn{2}{c}{Paired differences [95\% CI]} \\
\cmidrule(lr){2-5}\cmidrule(lr){6-7}
Setting & $C_{00}$ & $C_{10}$ & $C_{01}$ & $C_{11}$ & $C_{11}-C_{10}$ & $C_{11}-C_{01}$ \\
\midrule
M / 0 / 7 & $-1.1686$ & $-1.7487$ & $-1.1672$ & $-1.7552$ & $-0.0065$ $[-0.0112, -0.0018]$ & $-0.5880$ $[-0.6201, -0.5560]$ \\
M / 0 / 14 & $-1.4704$ & $-1.8690$ & $-1.4709$ & $-1.8698$ & $-0.0008$ $[-0.0050, 0.0035]$ & $-0.3989$ $[-0.4273, -0.3721]$ \\
M / 1 / 7 & $-1.2431$ & $-1.7101$ & $-1.2009$ & $-1.7804$ & $-0.0703$ $[-0.0801, -0.0603]$ & $-0.5794$ $[-0.6119, -0.5471]$ \\
M / 1 / 14 & $-1.4725$ & $-1.8960$ & $-1.4670$ & $-1.9039$ & $-0.0079$ $[-0.0110, -0.0049]$ & $-0.4369$ $[-0.4659, -0.4099]$ \\
L / 0 / 7 & $-1.3803$ & $-1.6929$ & $-1.4145$ & $-1.7190$ & $-0.0261$ $[-0.0356, -0.0164]$ & $-0.3045$ $[-0.3280, -0.2816]$ \\
L / 0 / 14 & $-1.6070$ & $-1.8832$ & $-1.6154$ & $-1.8893$ & $-0.0061$ $[-0.0105, -0.0017]$ & $-0.2739$ $[-0.2943, -0.2536]$ \\
L / 1 / 7 & $-1.4299$ & $-1.6590$ & $-1.4371$ & $-1.7548$ & $-0.0958$ $[-0.1101, -0.0821]$ & $-0.3176$ $[-0.3395, -0.2962]$ \\
L / 1 / 14 & $-1.6193$ & $-1.9002$ & $-1.6226$ & $-1.8976$ & $0.0027$ $[-0.0036, 0.0090]$ & $-0.2750$ $[-0.2970, -0.2532]$ \\
\bottomrule
\end{tabular}
\endgroup
\end{table*}
\begin{figure*}[!tp]
  \centering
  \includegraphics[width=\textwidth]{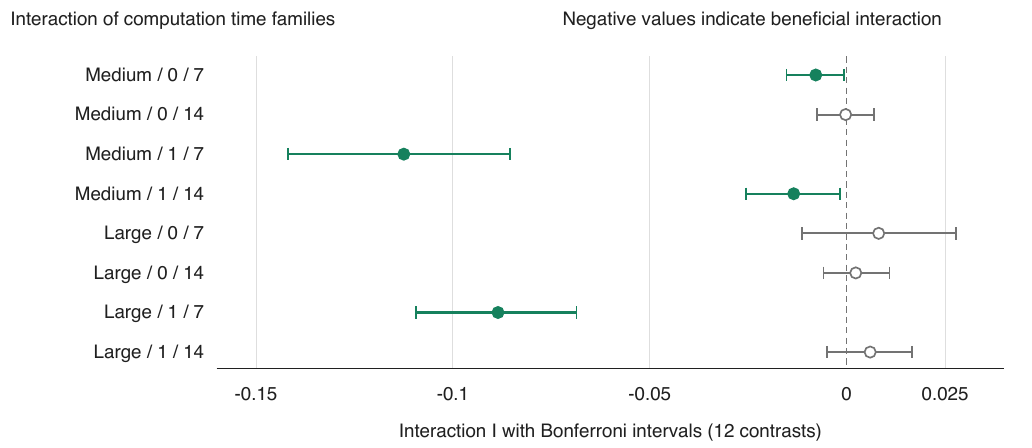}
  \caption{\textbf{Interaction between frozen full- and cached-computation time families.}
  Markers show the within-order interaction $I=C_{11}-C_{10}-C_{01}+C_{00}$
  evaluated on 100 DrawBench prompts with two initial noises per prompt
  (order 0/1 denotes zero-/first-order caching), where negative values
  indicate a beneficial super-additive interaction. Error bars denote
  99.5833\% Bonferroni-corrected confidence intervals across the 12-contrast
  family (eight interactions and four between-order differences); filled
  markers indicate intervals strictly below zero, whereas open markers span
  zero.}
  \label{fig:analysis-family-interaction}
\end{figure*}

After 12-comparison Bonferroni correction, confidence intervals for $I$ lie
strictly below zero in four settings---SD3 Medium zero-order $K=7$, SD3
Medium first-order $K=7$ and $K=14$, and SD3.5 Large first-order $K=7$---while
the remaining four span zero (\cref{fig:analysis-family-interaction}). The 68
prompts outside the 32-prompt diagnostic subset reproduce all four negative
interactions at unadjusted 95\% confidence.

At $K=7$, interactions are significantly stronger under first-order caching,
with $D=-0.1046$ (corrected CI $[-0.1360,-0.0750]$) on SD3 Medium and
$D=-0.0967$ ($[-0.1259,-0.0685]$) on SD3.5 Large. At $K=14$, SD3 Medium's
unadjusted 95\% interval lies below zero, whereas SD3.5 Large's contrast is
inconclusive.

Overall, Joint ($11$) achieves the lowest mean cost in seven of the eight
settings and outperforms both mixed arms in six at unadjusted 95\% confidence
(\cref{tab:analysis-family}). Only on SD3.5 Large with first-order caching at
$K=14$ does arm $10$ attain a slightly lower mean, with the interval for
$C_{11}-C_{10}$ spanning zero. Taken together, these tight-budget results
underscore the value of co-adapting both time families for first-order
forecasting, whose extrapolation depends directly on cross-family spacing.

\subsection{First-order history geometry}
\label{app:analysis-history}

For a cached step $t_i<t_b$ with two preceding full-computation noise levels
$t_a>t_b$, we characterize the local history geometry via the extrapolation
distance $d$, history spacing $g$, extrapolation ratio $\rho$, and Taylor
remainder scale $\kappa$:
\begin{equation}
 d=t_b-t_i,\quad g=t_a-t_b,\quad
 \rho=\frac{d}{g},\quad \kappa=\tfrac12d(d+g).
\label{eq:app_history_geometry}
\end{equation}
Here, $\rho$ governs the linear weights in
$\widehat F_i=(1+\rho)F_b-\rho F_a$ and can amplify prediction variance,
while $\kappa$ scales the leading second-derivative remainder along smooth
trajectories. At fixed $d$, narrowing $g$ increases $\rho$ while reducing
$\kappa$.

\begin{table*}[!tp]
\centering
\caption{\textbf{First-order history geometry and local error.}
Mean prompt-wise Pearson correlations of $\log_{10} r_i$ with $\log_{10}\rho$
and $\log_{10}\kappa$ across 32 DrawBench prompts (two initial noises per
prompt; arms $00$/$11$ denote Uniform/Joint). Confidence intervals apply the
32-test Bonferroni correction (99.84375\%). Linear trend removal subtracts
each variable's least-squares linear fit against noise level $t_i$
(\cref{app:analysis-history}).}
\label{tab:analysis-history}
\begingroup
\fontsize{9}{10.8}\selectfont
\setlength{\tabcolsep}{5pt}
\renewcommand{\arraystretch}{0.98}
\begin{tabular}{lccr@{\hspace{16pt}}r}
\toprule
Model / $K$ & Arm & Predictor & Direct [CI] & After linear trend removal [CI] \\
\midrule
\multirow{4}{*}{Medium / 7} & 00 & $\rho$ & 0.7658 $[0.7201, 0.8119]$ & 0.8963 $[0.8749, 0.9170]$ \\
 & 00 & $\kappa$ & 0.7305 $[0.6833, 0.7772]$ & 0.8205 $[0.7856, 0.8528]$ \\
 & 11 & $\rho$ & 0.9445 $[0.9297, 0.9571]$ & 0.9325 $[0.9158, 0.9478]$ \\
 & 11 & $\kappa$ & 0.7321 $[0.6898, 0.7727]$ & 0.7656 $[0.7354, 0.7949]$ \\
\addlinespace[2pt]
\multirow{4}{*}{Medium / 14} & 00 & $\rho$ & 0.2071 $[0.1721, 0.2534]$ & 0.2595 $[0.2064, 0.3180]$ \\
 & 00 & $\kappa$ & 0.0123 $[-0.0356, 0.0805]$ & 0.0193 $[-0.0466, 0.1026]$ \\
 & 11 & $\rho$ & 0.7323 $[0.6712, 0.7895]$ & 0.9103 $[0.8895, 0.9274]$ \\
 & 11 & $\kappa$ & 0.7072 $[0.6437, 0.7661]$ & 0.8198 $[0.7843, 0.8502]$ \\
\addlinespace[2pt]
\multirow{4}{*}{Large / 7} & 00 & $\rho$ & 0.7895 $[0.7558, 0.8221]$ & 0.8706 $[0.8522, 0.8867]$ \\
 & 00 & $\kappa$ & 0.7435 $[0.7081, 0.7766]$ & 0.7959 $[0.7638, 0.8239]$ \\
 & 11 & $\rho$ & 0.9565 $[0.9294, 0.9738]$ & 0.9998 $[0.9996, 0.9999]$ \\
 & 11 & $\kappa$ & 0.9716 $[0.9486, 0.9858]$ & 0.9998 $[0.9996, 0.9999]$ \\
\addlinespace[2pt]
\multirow{4}{*}{Large / 14} & 00 & $\rho$ & 0.1703 $[0.1382, 0.2072]$ & 0.1780 $[0.1359, 0.2274]$ \\
 & 00 & $\kappa$ & $-0.0724$ $[-0.1086, -0.0332]$ & $-0.0815$ $[-0.1228, -0.0352]$ \\
 & 11 & $\rho$ & 0.5982 $[0.5309, 0.6649]$ & 0.6031 $[0.5350, 0.6709]$ \\
 & 11 & $\kappa$ & 0.4886 $[0.4169, 0.5594]$ & 0.5708 $[0.5000, 0.6410]$ \\
\bottomrule
\end{tabular}
\endgroup
\end{table*}
\begin{figure*}[!tp]
  \input{figures/analysis_history_geometry}
\end{figure*}

Across the four first-order settings (512 trajectories), 126 two-history
positions yield 8,064 records over 64 prompt--noise inputs (excluding 896
single-history records that fall back to zero-order reuse). For each prompt,
we average $r_i$ across the two noises per position, compute Pearson
correlations of $\log_{10}r_i$ with $\log_{10}\rho$ and $\log_{10}\kappa$
across positions, and average the resulting coefficients across all 32
prompts. Detrended correlations remove the least-squares linear trend in
$t_i$ from both variables beforehand.

\Needspace{6\baselineskip}
After Bonferroni correction, all four Joint schedules exhibit strictly
positive direct and detrended correlations for both predictors
(\cref{tab:analysis-history}). Across Uniform and Joint combined, 28 of 32
tests yield intervals above zero; the only four neutral or negative
associations occur for $\kappa$ under Uniform at $K=14$
(\cref{fig:analysis-history-geometry}).

\Needspace{6\baselineskip}
On SD3.5 Large with first-order caching at $K=7$, 12 of Joint's 18
two-history positions have $\rho>1+10^{-6}$ (minimum $g\approx 0.0388$,
maximum $\rho\approx 1.65$). Despite these larger extrapolation ratios, Joint
reduces total $H_R$ by 66.0\%, $A$ by 40.1\%, and $C$ by $0.3395$ relative to
Uniform (\cref{tab:analysis-local-error}).
\par

\twocolumn[{%
\section{Additional qualitative comparisons}
\label{app:sd3-qualitative}

\Cref{fig:sd35-large-qualitative} extends Figure~4 of the main paper to
SD3.5 Large~\citep{stabilityai2024sd35} under first-order caching at
$512\times512$ resolution ($N=28$, target $K=7$), following the generation
protocol and scheduling adapters in \cref{app:settings}.
\Cref{fig:flux-qualitative} further presents a standalone zero-order caching
comparison on FLUX.1-dev~\citep{blackforestlabs-flux1-dev-model-card}
at $1024\times1024$ resolution ($N=50$, target
$K=13$) across four DrawBench prompts~\citep{NEURIPS2022_ec795aea} probing
color binding and material texture, where offline policies (ReCache and
Joint) are trained and online adaptive thresholds (DiCache and SeaCache) are
calibrated specifically for this setting. In both figures, the column labels
$\mathrm{NFE}=7$ and $\mathrm{NFE}=13$ denote exact full-computation budgets
for static policies and target mean budgets for online adaptive methods.

\medskip
\noindent\begin{minipage}{\textwidth}
  \captionsetup{type=figure,skip=5pt}
  \centering
  \includegraphics[width=\textwidth]{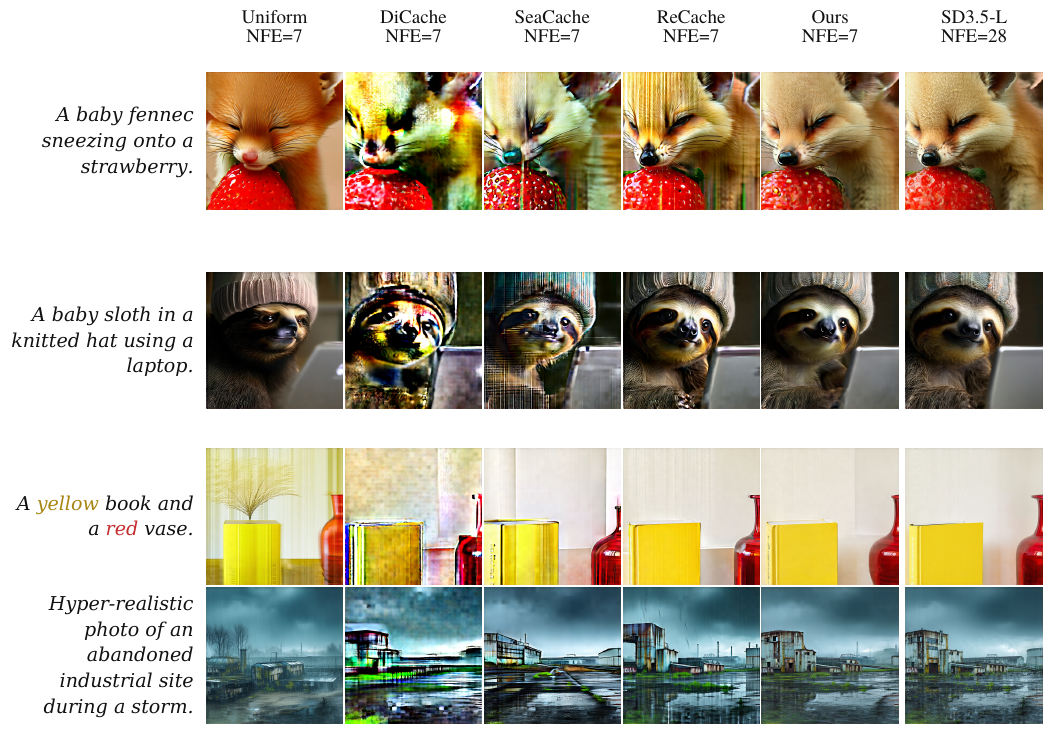}\par
  \caption{\textbf{Qualitative comparisons under first-order caching on SD3.5 Large.}
  Generated at $512\times512$ resolution ($N=28$) following the protocol and
  scheduling adapters in \cref{app:settings}; each row shares the same prompt
  and initial noise across methods (the first two prompts are abbreviated).
  Uniform, ReCache, and Ours (Joint) execute exactly $K=7$ full computations
  (NFE), DiCache and SeaCache target an average of $K=7$, and the rightmost
  column shows the 28-step full-inference reference.}
  \label{fig:sd35-large-qualitative}
\end{minipage}\par
}]

\onecolumn
\noindent\begin{minipage}{\textwidth}
  \captionsetup{type=figure,skip=6pt}
  \centering
  \includegraphics[width=\textwidth]{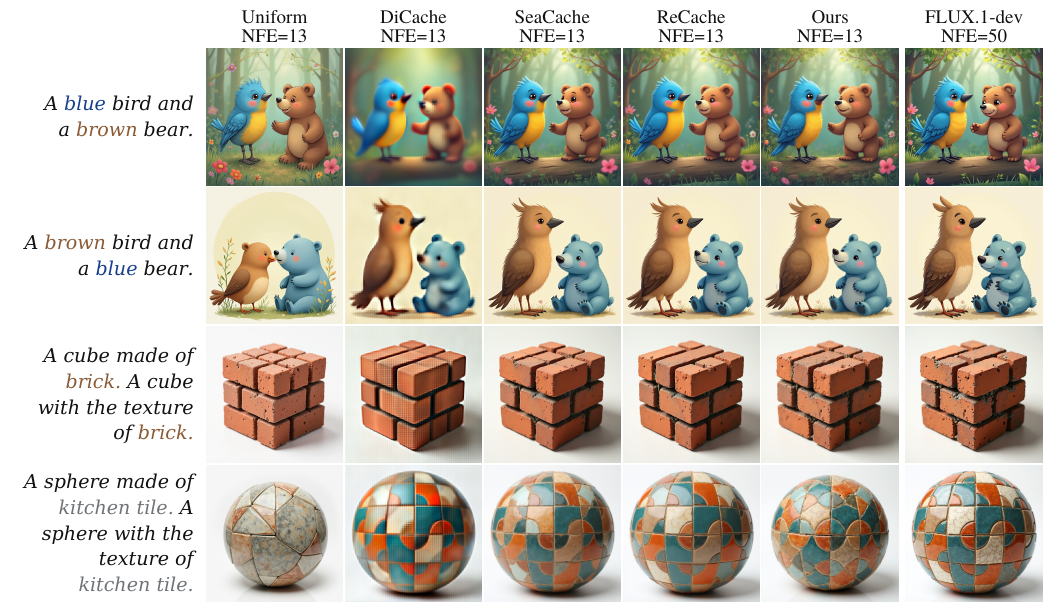}\par
  \caption{\textbf{Qualitative comparisons under zero-order caching on FLUX.1-dev.}
  Four DrawBench prompts~\citep{NEURIPS2022_ec795aea} probing color binding
  and material texture are generated at $1024\times1024$ resolution
  ($N=50$), with each row displaying the full prompt and sharing the same
  initial noise across methods. The five caching policies target $K=13$ full
  computations (NFE)---with realized counts varying per sample for online
  adaptive methods---whereas the reference shows 50-step full inference.}
  \label{fig:flux-qualitative}
\end{minipage}\par

\section{Extended scheduling comparisons}
\label{app:extended-comparisons}

\subsection{FLUX.1-dev at a fixed full-computation budget}
\label{app:extended-flux}
We evaluate both caching orders at the resolution and budget of \cref{fig:flux-qualitative} on 1,000 paired test images.

\smallskip
\noindent\begin{minipage}{\linewidth}
\captionsetup{type=table,skip=5pt}
\centering
\caption{\textbf{FLUX.1-dev schedules at $N=50$, $K=13$.} Static schedules use exact $K$. DiCache and SeaCache target mean $K$. Fidelity uses native 50-step full inference (PSNR in dB). Bold and underline mark the best and second-best scores.}
\label{tab:extended-flux-budget}
\fontsize{9}{10.8}\selectfont
\setlength{\tabcolsep}{7pt}
\renewcommand{\arraystretch}{1.06}
\textbf{(a) Zero-order}\par\vspace{2pt}
\begin{tabular}{l*{6}{r}}
\toprule
\textbf{Scheduling policy} & \textbf{LPIPS\boldmath$\downarrow$} & \textbf{SSIM\boldmath$\uparrow$} & \textbf{PSNR\boldmath$\uparrow$} & \textbf{ImageReward\boldmath$\uparrow$} & \textbf{PickScore\boldmath$\uparrow$} & \textbf{HPSv2\boldmath$\uparrow$} \\
\midrule
Uniform & 0.4500 & 0.6447 & 14.85 & 0.9495 & 22.7036 & 0.292862 \\
DiCache & 0.4588 & 0.6705 & 19.20 & 0.3356 & 21.5142 & 0.270965 \\
SeaCache & 0.2145 & 0.7992 & 21.08 & 0.9774 & 22.7701 & 0.295981 \\
ReCache & \textbf{0.1834} & 0.8172 & \textbf{22.77} & 0.9811 & 22.7724 & 0.295865 \\
\rowcolor{mainResultCyan}Joint (ALR) & 0.1946 & \underline{0.8201} & 22.15 & \underline{0.9863} & \textbf{22.8132} & \underline{0.296030} \\
\rowcolor{mainResultCyan}\textbf{Joint (span/start)} & \underline{0.1885} & \textbf{0.8202} & \underline{22.34} & \textbf{0.9879} & \underline{22.8015} & \textbf{0.296098} \\
\bottomrule
\end{tabular}\par
\vspace{6pt}
\textbf{(b) First-order}\par\vspace{2pt}
\begin{tabular}{l*{6}{r}}
\toprule
\textbf{Scheduling policy} & \textbf{LPIPS\boldmath$\downarrow$} & \textbf{SSIM\boldmath$\uparrow$} & \textbf{PSNR\boldmath$\uparrow$} & \textbf{ImageReward\boldmath$\uparrow$} & \textbf{PickScore\boldmath$\uparrow$} & \textbf{HPSv2\boldmath$\uparrow$} \\
\midrule
Uniform & 0.4339 & 0.6497 & 15.00 & 0.9795 & 22.7954 & 0.294326 \\
DiCache & 0.8178 & 0.3093 & 12.52 & -0.1818 & 20.6251 & 0.264978 \\
SeaCache & 0.1750 & 0.8254 & 21.50 & \textbf{1.0123} & \textbf{22.8986} & \underline{0.297170} \\
ReCache & \underline{0.1628} & \underline{0.8301} & \underline{22.39} & 1.0029 & 22.8563 & 0.295737 \\
\rowcolor{mainResultCyan}Joint (ALR) & 0.1914 & 0.8123 & 20.99 & \underline{1.0107} & 22.8859 & 0.297161 \\
\rowcolor{mainResultCyan}\textbf{Joint (span/start)} & \textbf{0.1531} & \textbf{0.8447} & \textbf{22.58} & 1.0091 & \underline{22.8920} & \textbf{0.297439} \\
\bottomrule
\end{tabular}\par
\end{minipage}
\par
\clearpage

We use $1024\times1024$ images, BF16, guidance 3.5 and five noise seeds for each of 200 held-out prompts.
We train both Joint encodings from \cref{app:training_configs} for each backend with policy seed 0. Span/start uses span--shape
for full computations and start--shape for cached computations.
Relative to ReCache, span/start lowers first-order LPIPS from 0.1628 to 0.1531. For zero-order caching,
ReCache has lower LPIPS (0.1834 vs.\ 0.1885) and span/start has higher SSIM (0.8202 vs.\ 0.8172).

\subsection{FLUX.1-dev at comparable inference latency}
\label{app:extended-flux-latency}
For each span/start schedule in \cref{tab:extended-flux-budget}, we select the
native-shift Euler step count closest in mean validation latency. A second
Euler grid spaces $\sigma$ evenly from 1 to 0 at the same step count.

\smallskip
\noindent\begin{minipage}{\linewidth}
\captionsetup{type=table,skip=5pt}
\centering
\caption{\textbf{Few-step full Euler inference at comparable FLUX.1-dev latency.} Each Euler update runs the full network. Time differences are relative to Joint. We use the same 1,000 paired inputs as \cref{tab:extended-flux-budget} and the native 50-step fidelity reference. Bold marks each group's best score (PSNR in dB).}
\label{tab:extended-flux-euler}
\fontsize{9}{10.8}\selectfont
\setlength{\tabcolsep}{2.35pt}
\renewcommand{\arraystretch}{1.12}
\begin{tabular}{lcccc*{6}{r}}
\toprule
\textbf{Sampling schedule} & \textbf{Updates/full} & \textbf{s/image} & \textbf{\boldmath$\Delta t$ (s)} & \textbf{\boldmath$\Delta t$ (\%)} & \textbf{LPIPS\boldmath$\downarrow$} & \textbf{SSIM\boldmath$\uparrow$} & \textbf{PSNR\boldmath$\uparrow$} & \textbf{ImageReward\boldmath$\uparrow$} & \textbf{PickScore\boldmath$\uparrow$} & \textbf{HPSv2\boldmath$\uparrow$} \\
\midrule
\multicolumn{11}{l}{\textbf{(a) Matched to zero-order Joint}} \\
\rowcolor{mainResultCyan}\textbf{Joint (span/start)} & 50/13 & 1.9078 & --- & --- & \textbf{0.1885} & \textbf{0.8202} & \textbf{22.34} & \textbf{0.9879} & \textbf{22.8015} & \textbf{0.296098} \\
Euler, native shift & 19/19 & 1.9449 & $+0.0370$ & $+1.94$ & 0.3783 & 0.6913 & 16.09 & 0.9820 & 22.7967 & 0.294831 \\
Euler, uniform $\sigma$ & 19/19 & 1.9443 & $+0.0365$ & $+1.91$ & 0.5232 & 0.6005 & 13.69 & 0.7494 & 22.5028 & 0.287560 \\
\midrule
\multicolumn{11}{l}{\textbf{(b) Matched to first-order Joint}} \\
\rowcolor{mainResultCyan}\textbf{Joint (span/start)} & 50/13 & 3.0356 & --- & --- & \textbf{0.1531} & \textbf{0.8447} & \textbf{22.58} & \textbf{1.0091} & \textbf{22.8920} & 0.297439 \\
Euler, native shift & 30/30 & 3.0173 & $-0.0182$ & $-0.60$ & 0.2408 & 0.7857 & 19.40 & 1.0046 & 22.8509 & \textbf{0.297456} \\
Euler, uniform $\sigma$ & 30/30 & 3.0169 & $-0.0186$ & $-0.61$ & 0.4729 & 0.6310 & 14.31 & 0.8966 & 22.7041 & 0.291883 \\
\bottomrule
\end{tabular}\par
\end{minipage}
\par
\smallskip
We time denoising and VAE decoding for all methods on the same GPU at batch size~1, using cached conditioning.
After three warmups per method, we interleave 60 timed runs per method over 20 validation inputs and three rounds.
Against 19- and 30-step native Euler, the respective LPIPS reductions are 0.1898 and 0.0877.
Their paired 95\% bootstrap intervals are $[0.1792,0.2006]$ and $[0.0792,0.0963]$
(20,000 prompt resamples, keeping each prompt's five seeds together).
First-order HPSv2 differs by 0.000017 in favor of native Euler.

\subsection{Additional schedulers on SD3}
\label{app:extended-sd3}
We compare Joint with BudCache~\citep{lei2026budgetconstrained} and ERTACache~\citep{peng2026ertacache}
on SD3 Medium (Table~\ref{tab:extended-sd3-medium}) and SD3.5 Large (Table~\ref{tab:extended-sd35-large}).
BudCache's first stage optimizes cache refresh decisions under a fixed compute budget, and its second stage
optimizes the sampling times with those decisions held fixed.

\smallskip
\noindent\begin{minipage}{\linewidth}
\captionsetup{type=table,skip=5pt}
\centering
\caption{\textbf{Additional scheduling comparisons on SD3 Medium.} Joint scores are from Table~1 of the main paper. Bold and underline denote the best and second-best values at displayed precision (PSNR in dB).}
\label{tab:extended-sd3-medium}
\fontsize{9}{10.8}\selectfont
\setlength{\tabcolsep}{2.85pt}
\renewcommand{\arraystretch}{1.01}
\textbf{(a) Zero-order}\par\vspace{2pt}
\begin{tabular}{l*{12}{c}}
\toprule
\multirow{2}{*}{\textbf{Scheduling policy}} & \multicolumn{2}{c}{\textbf{LPIPS\boldmath$\downarrow$}} & \multicolumn{2}{c}{\textbf{SSIM\boldmath$\uparrow$}} & \multicolumn{2}{c}{\textbf{PSNR\boldmath$\uparrow$}} & \multicolumn{2}{c}{\textbf{ImageReward\boldmath$\uparrow$}} & \multicolumn{2}{c}{\textbf{PickScore\boldmath$\uparrow$}} & \multicolumn{2}{c}{\textbf{HPSv2\boldmath$\uparrow$}} \\
\cmidrule(lr){2-3}\cmidrule(lr){4-5}\cmidrule(lr){6-7}\cmidrule(lr){8-9}\cmidrule(lr){10-11}\cmidrule(lr){12-13}
& \makebox[26pt]{\textbf{\boldmath$K{=}7$}} & \makebox[26pt]{\textbf{\boldmath$K{=}14$}} & \makebox[26pt]{\textbf{\boldmath$K{=}7$}} & \makebox[26pt]{\textbf{\boldmath$K{=}14$}} & \makebox[26pt]{\textbf{\boldmath$K{=}7$}} & \makebox[26pt]{\textbf{\boldmath$K{=}14$}} & \makebox[26pt]{\textbf{\boldmath$K{=}7$}} & \makebox[26pt]{\textbf{\boldmath$K{=}14$}} & \makebox[26pt]{\textbf{\boldmath$K{=}7$}} & \makebox[26pt]{\textbf{\boldmath$K{=}14$}} & \makebox[26pt]{\textbf{\boldmath$K{=}7$}} & \makebox[26pt]{\textbf{\boldmath$K{=}14$}} \\
\midrule
BudCache & \textbf{0.2670} & \underline{0.1433} & \underline{0.6194} & \underline{0.8237} & \underline{17.34} & \underline{21.07} & \textbf{0.318} & \textbf{0.557} & \underline{21.04} & \textbf{21.59} & \underline{0.2739} & \textbf{0.2812} \\
ERTACache & 0.4573 & 0.1484 & 0.5156 & 0.8202 & 13.38 & 20.90 & -0.188 & \underline{0.550} & 20.57 & \underline{21.56} & 0.2661 & 0.2807 \\
\rowcolor{mainResultCyan}\textbf{Joint} & \underline{0.2678} & \textbf{0.1326} & \textbf{0.6746} & \textbf{0.8307} & \textbf{17.38} & \textbf{21.29} & \underline{0.298} & 0.533 & \textbf{21.17} & 21.54 & \textbf{0.2759} & \underline{0.2808} \\
\bottomrule
\end{tabular}\par
\vspace{4pt}
\textbf{(b) First-order}\par\vspace{2pt}
\begin{tabular}{l*{12}{c}}
\toprule
\multirow{2}{*}{\textbf{Scheduling policy}} & \multicolumn{2}{c}{\textbf{LPIPS\boldmath$\downarrow$}} & \multicolumn{2}{c}{\textbf{SSIM\boldmath$\uparrow$}} & \multicolumn{2}{c}{\textbf{PSNR\boldmath$\uparrow$}} & \multicolumn{2}{c}{\textbf{ImageReward\boldmath$\uparrow$}} & \multicolumn{2}{c}{\textbf{PickScore\boldmath$\uparrow$}} & \multicolumn{2}{c}{\textbf{HPSv2\boldmath$\uparrow$}} \\
\cmidrule(lr){2-3}\cmidrule(lr){4-5}\cmidrule(lr){6-7}\cmidrule(lr){8-9}\cmidrule(lr){10-11}\cmidrule(lr){12-13}
& \makebox[26pt]{\textbf{\boldmath$K{=}7$}} & \makebox[26pt]{\textbf{\boldmath$K{=}14$}} & \makebox[26pt]{\textbf{\boldmath$K{=}7$}} & \makebox[26pt]{\textbf{\boldmath$K{=}14$}} & \makebox[26pt]{\textbf{\boldmath$K{=}7$}} & \makebox[26pt]{\textbf{\boldmath$K{=}14$}} & \makebox[26pt]{\textbf{\boldmath$K{=}7$}} & \makebox[26pt]{\textbf{\boldmath$K{=}14$}} & \makebox[26pt]{\textbf{\boldmath$K{=}7$}} & \makebox[26pt]{\textbf{\boldmath$K{=}14$}} & \makebox[26pt]{\textbf{\boldmath$K{=}7$}} & \makebox[26pt]{\textbf{\boldmath$K{=}14$}} \\
\midrule
BudCache & \underline{0.3519} & \textbf{0.1290} & \underline{0.5769} & \textbf{0.8462} & \underline{14.82} & \textbf{21.40} & \underline{0.066} & \textbf{0.572} & \underline{20.63} & \textbf{21.62} & \underline{0.2688} & \textbf{0.2815} \\
ERTACache & 0.5036 & 0.1334 & 0.4709 & 0.8389 & 12.11 & \underline{21.25} & -0.180 & \underline{0.567} & 20.42 & \underline{21.60} & 0.2649 & \underline{0.2814} \\
\rowcolor{mainResultCyan}\textbf{Joint} & \textbf{0.2395} & \underline{0.1328} & \textbf{0.7097} & \underline{0.8412} & \textbf{17.62} & 21.15 & \textbf{0.350} & 0.548 & \textbf{21.18} & \underline{21.60} & \textbf{0.2721} & 0.2810 \\
\bottomrule
\end{tabular}\par
\end{minipage}
\par

\twocolumn[{%
\noindent\begin{minipage}{\linewidth}
\captionsetup{type=table,skip=5pt}
\centering
\caption{\textbf{Additional scheduling comparisons on SD3.5 Large.} Joint scores are from Table~2 of the main paper. Bold and underline denote the best and second-best values at displayed precision (PSNR in dB).}
\label{tab:extended-sd35-large}
\fontsize{9}{10.8}\selectfont
\setlength{\tabcolsep}{2.85pt}
\renewcommand{\arraystretch}{1.01}
\textbf{(a) Zero-order}\par\vspace{2pt}
\begin{tabular}{l*{12}{c}}
\toprule
\multirow{2}{*}{\textbf{Scheduling policy}} & \multicolumn{2}{c}{\textbf{LPIPS\boldmath$\downarrow$}} & \multicolumn{2}{c}{\textbf{SSIM\boldmath$\uparrow$}} & \multicolumn{2}{c}{\textbf{PSNR\boldmath$\uparrow$}} & \multicolumn{2}{c}{\textbf{ImageReward\boldmath$\uparrow$}} & \multicolumn{2}{c}{\textbf{PickScore\boldmath$\uparrow$}} & \multicolumn{2}{c}{\textbf{HPSv2\boldmath$\uparrow$}} \\
\cmidrule(lr){2-3}\cmidrule(lr){4-5}\cmidrule(lr){6-7}\cmidrule(lr){8-9}\cmidrule(lr){10-11}\cmidrule(lr){12-13}
& \makebox[26pt]{\textbf{\boldmath$K{=}7$}} & \makebox[26pt]{\textbf{\boldmath$K{=}14$}} & \makebox[26pt]{\textbf{\boldmath$K{=}7$}} & \makebox[26pt]{\textbf{\boldmath$K{=}14$}} & \makebox[26pt]{\textbf{\boldmath$K{=}7$}} & \makebox[26pt]{\textbf{\boldmath$K{=}14$}} & \makebox[26pt]{\textbf{\boldmath$K{=}7$}} & \makebox[26pt]{\textbf{\boldmath$K{=}14$}} & \makebox[26pt]{\textbf{\boldmath$K{=}7$}} & \makebox[26pt]{\textbf{\boldmath$K{=}14$}} & \makebox[26pt]{\textbf{\boldmath$K{=}7$}} & \makebox[26pt]{\textbf{\boldmath$K{=}14$}} \\
\midrule
BudCache & \underline{0.3019} & \textbf{0.1286} & \underline{0.6378} & \textbf{0.8649} & \underline{19.77} & \textbf{24.88} & \underline{0.273} & \textbf{0.543} & \underline{21.32} & \textbf{21.82} & \textbf{0.2731} & \underline{0.2797} \\
ERTACache & 0.3938 & 0.1373 & 0.5919 & 0.8462 & 17.88 & \underline{24.69} & 0.065 & \underline{0.534} & 21.14 & \underline{21.79} & 0.2687 & \textbf{0.2798} \\
\rowcolor{mainResultCyan}\textbf{Joint} & \textbf{0.2916} & \underline{0.1294} & \textbf{0.7024} & \underline{0.8633} & \textbf{19.78} & 24.12 & \textbf{0.323} & \textbf{0.543} & \textbf{21.47} & \textbf{21.82} & \underline{0.2728} & \underline{0.2797} \\
\bottomrule
\end{tabular}\par
\vspace{4pt}
\textbf{(b) First-order}\par\vspace{2pt}
\begin{tabular}{l*{12}{c}}
\toprule
\multirow{2}{*}{\textbf{Scheduling policy}} & \multicolumn{2}{c}{\textbf{LPIPS\boldmath$\downarrow$}} & \multicolumn{2}{c}{\textbf{SSIM\boldmath$\uparrow$}} & \multicolumn{2}{c}{\textbf{PSNR\boldmath$\uparrow$}} & \multicolumn{2}{c}{\textbf{ImageReward\boldmath$\uparrow$}} & \multicolumn{2}{c}{\textbf{PickScore\boldmath$\uparrow$}} & \multicolumn{2}{c}{\textbf{HPSv2\boldmath$\uparrow$}} \\
\cmidrule(lr){2-3}\cmidrule(lr){4-5}\cmidrule(lr){6-7}\cmidrule(lr){8-9}\cmidrule(lr){10-11}\cmidrule(lr){12-13}
& \makebox[26pt]{\textbf{\boldmath$K{=}7$}} & \makebox[26pt]{\textbf{\boldmath$K{=}14$}} & \makebox[26pt]{\textbf{\boldmath$K{=}7$}} & \makebox[26pt]{\textbf{\boldmath$K{=}14$}} & \makebox[26pt]{\textbf{\boldmath$K{=}7$}} & \makebox[26pt]{\textbf{\boldmath$K{=}14$}} & \makebox[26pt]{\textbf{\boldmath$K{=}7$}} & \makebox[26pt]{\textbf{\boldmath$K{=}14$}} & \makebox[26pt]{\textbf{\boldmath$K{=}7$}} & \makebox[26pt]{\textbf{\boldmath$K{=}14$}} & \makebox[26pt]{\textbf{\boldmath$K{=}7$}} & \makebox[26pt]{\textbf{\boldmath$K{=}14$}} \\
\midrule
BudCache & \underline{0.3142} & 0.1303 & \underline{0.6619} & 0.8632 & \underline{17.69} & 23.75 & \underline{0.266} & 0.557 & \underline{21.17} & 21.79 & \textbf{0.2744} & 0.2799 \\
ERTACache & 0.5088 & \underline{0.1053} & 0.5103 & \underline{0.8874} & 14.11 & \textbf{25.52} & 0.124 & \underline{0.577} & 21.04 & \underline{21.83} & 0.2716 & \textbf{0.2807} \\
\rowcolor{mainResultCyan}\textbf{Joint} & \textbf{0.2779} & \textbf{0.1018} & \textbf{0.6845} & \textbf{0.8917} & \textbf{18.91} & \underline{25.46} & \textbf{0.300} & \textbf{0.581} & \textbf{21.24} & \textbf{21.84} & \underline{0.2734} & \underline{0.2801} \\
\bottomrule
\end{tabular}\par
\end{minipage}
\par
\bigskip
}]

\begingroup
\small
\setlength{\baselineskip}{9.6pt}
\let\url\nolinkurl
\bibliographystyle{preprint-fullnames}
\bibliography{references,references-supplementary}
\endgroup